\documentclass{article}

\PassOptionsToPackage{numbers, compress}{natbib}

\usepackage[main, final]{neurips_2026}

\usepackage[utf8]{inputenc} 
\usepackage[T1]{fontenc}    
\usepackage{hyperref}       
\usepackage{url}            
\usepackage{booktabs}       
\usepackage{amsfonts}       
\usepackage{nicefrac}       
\usepackage{microtype}      
\usepackage{xcolor}         

\usepackage{graphicx}
\usepackage{subcaption}
\usepackage{tocloft}
\usepackage{etoc}

\usepackage{pifont}
\usepackage{amsmath}
\usepackage{amssymb}
\usepackage{mathtools}
\usepackage{amsthm}
\usepackage{enumitem}
\usepackage{wrapfig}

\usepackage{makecell}
\usepackage[table]{xcolor}
\definecolor{best2}{HTML}{C8E6C9}
\definecolor{best}{HTML}{BBDEFB}
\definecolor{grey}{HTML}{e0e0e0}

\title{ImagineAgent: What if Agents Could Imagine? \\ Reinforcing Open-Vocabulary HOI Comprehension through Generation}

\author{
\textbf{
Zhenlong Yuan\textsuperscript{1 *} \quad
Yue Wang\textsuperscript{2 *} \quad
Dapeng Zhang\textsuperscript{3 *} \quad
Kejin Cui\textsuperscript{4} \quad
}
\\[0.6em]
\textbf{
Rui Chen\textsuperscript{1}  \quad
Jing Tang\textsuperscript{1} \quad
Lei Sun\textsuperscript{1 \ensuremath{\ddagger}} \quad
Hongwei Yu\textsuperscript{1} \quad
}
\\[0.6em]
\textbf{
Chengxuan Qian\textsuperscript{1} \quad
Xiangxiang Chu\textsuperscript{1} \quad
Shuo Li\textsuperscript{5} \quad
Yuyin Zhou\textsuperscript{6 \textsection}
}
\\[1.0em]
\textsuperscript{1} AMAP, Alibaba Group \quad
\textsuperscript{2} Meta Platforms \quad
\textsuperscript{3} National University of Singapore \quad
\\[0.6em]
\textsuperscript{4} Independent Researcher \quad
\textsuperscript{5} Case Western Reserve University \quad
\textsuperscript{6} UC Santa Cruz
\\[1.0em]
\textsuperscript{*}~\textit{Equal contribution}\quad
\textsuperscript{\ensuremath{\ddagger}}~\textit{Project Lead}\quad
\textsuperscript{\textsection}~\textit{Corresponding Author}
}

\begin{document}

\maketitle

\begin{abstract}
Multimodal Large Language Models have shown promising capabilities in bridging visual and textual reasoning, yet their reasoning capabilities in Open-Vocabulary Human-Object Interaction (OV-HOI) are limited by cross-modal hallucinations and limited viewpoints of images. To address this, we propose \textbf{ImagineAgent}, an agentic framework that integrates cognitive mapping, tool-augmented reinforcement learning (RL), and generative world modeling for robust OV-HOI understanding.
Specifically, we first propose an innovative CoT dataset named hicodet-6K for supervised fine-tuning (SFT), which effectively bridges the perception-to-cognition gap by structuring perceived entities into interaction pairs for comprehensive predictions.
Subsequently, we develop a multimodal tool library integrating online retrieval, image cropping, and generative modeling, enabling the agent to dynamically augment reasoning with domain-specific tools to resolve visual-semantic ambiguities and hallucinations during inference.
Moreover, we incorporate a generative model to reconstruct alternative viewpoints, enabling the agent to “imagine” under limited viewpoints.
Finally, we propose a composite reward mechanism to jointly optimize prediction accuracy and tool efficiency. 
Evaluations on both SWIG-HOI and HICO-DET datasets demonstrate that our method achieves state-of-the-art performance while requiring merely 36.7\% of the training data compared to existing methods, validating our robustness, empirical effectiveness and efficiency.


\end{abstract}

\begin{figure} [h]
  \centering
  \includegraphics[width=\textwidth]{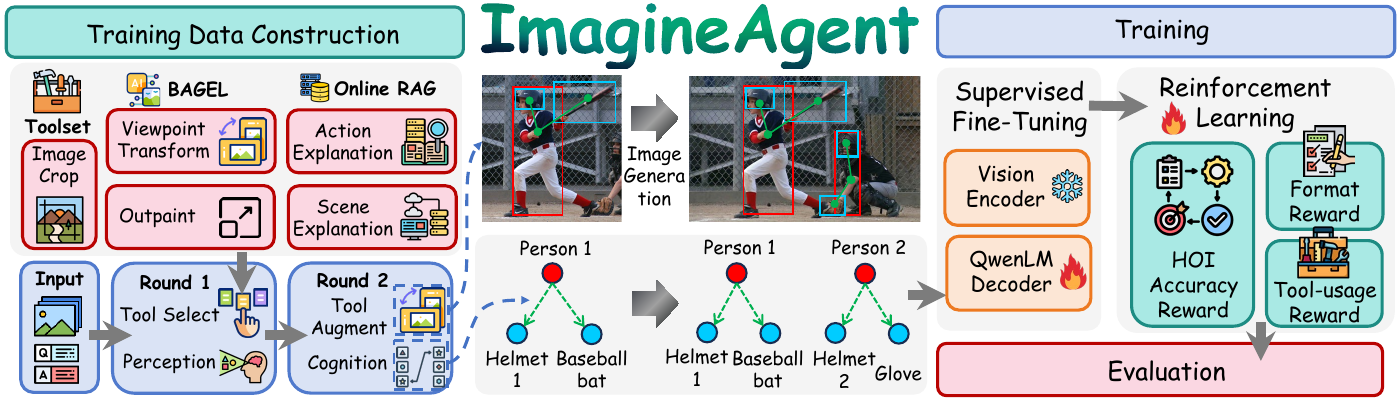}
\end{figure}

\section{Introduction}
Human-Object Interaction (HOI) detection~\cite{gupta2015visual} aims to identify and localize interacting human and object pairs within an image, while also recognizing their semantic relationships as \textit{<human, object, action>} triplets. This fine-grained understanding of human-centric activities is foundational for next-generation applications in areas such as robot manipulation, assistive technologies, and intelligent surveillance~\cite{liu2022interactiveness, kim2023relational, yang2024MPHOI}. To truly unlock this potential, a critical frontier is extending HOI detection beyond a predefined, closed set of categories to an open-vocabulary (OV) setting~\cite{wang2022_THID, lei2024CMD-SE, hu2025bilateral}. This capability would enable the system to generalize to detect any conceivable interaction, even those unseen during training~\cite{lei2024ez, kim2025locality}, thereby enabling true adaptability in the unpredictable real world.

\begin{figure*}[t]
  \begin{center}
    \centerline{\includegraphics[width=\textwidth]{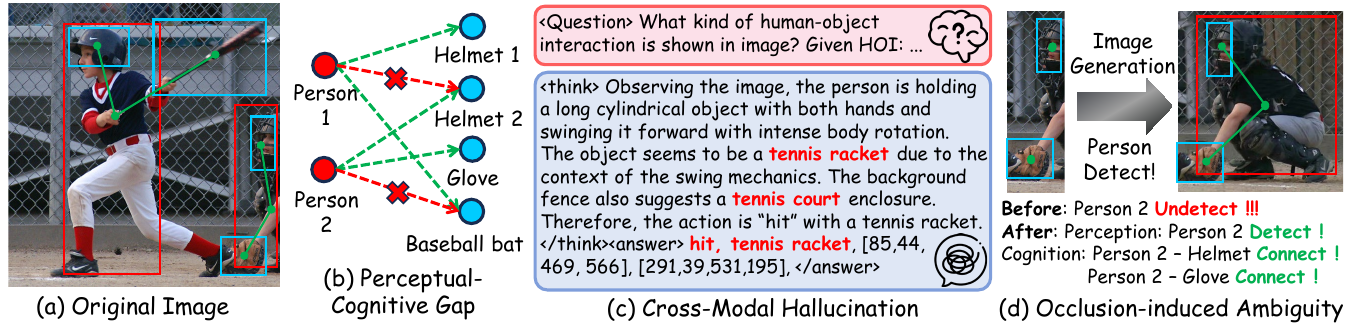}}
    \caption{
    \textbf{Three limitations in OV-HOI tasks.} 
    Given (a) an original image, MLLMs may suffer from (b) perceptual-cognitive gap, where they perceive individual entities but fail to form a coherent understanding, leading to flawed cognition (e.g., linking Person 1 to Helmet 2 and Glove they are not wearing).
    Another critical issue is (c) cross-modal hallucination, where models generate plausible-sounding but visually unsupported interactions by over-relying on textual priors (e.g., mistaking a baseball bat for a tennis racket). 
    Finally, (d) occlusion-induced ambiguity arises when key visual information is missing, causing the model fail to detect entities and their connections. 
    }
    \vspace{-0.25in}
    \label{abstract}
  \end{center}
\end{figure*}

Recent breakthroughs in Multimodal Large Language Models (MLLMs)~\cite{radford2021learning, hurst2024gpt, liu2024improved, comanici2025gemini, wang2025internvl3} have revolutionized both visual perception and reasoning. By leveraging the vast world knowledge embedded in their underlying language models and grounding it in visual evidence, MLLMs interpret nuanced visual details and their semantic implications with high fidelity. Models like Qwen-VL series~\cite{bai2025qwen2, bai2025qwen3} have demonstrated exceptional performance across a spectrum of challenging tasks, including visual dialogue, few-shot learning, and generating coherent narratives from visual inputs.





Despite their capabilities, MLLMs still exhibit three fundamental limitations when applied to OV-HOI tasks:
\ding{182} \textbf{Perceptual-Cognitive Gap}: As illustrated in Fig.~\ref{abstract} (b), while MLLMs detect all persons and objects individually, they often fail to establish semantically coherent physical relationships between them. This stems from a lack of structured cognitive mapping that connects visual perception to plausible interaction hypotheses, which may cause incomplete or fragmented reasoning despite accurate detection.
\ding{183} \textbf{Cross-Modal Hallucination}: As shown in Fig.~\ref{abstract} (c), MLLMs frequently generate plausible-sounding but visually unsupported interactions, driven by textual priors rather than visual cues. This cross-modal hallucination arises from imprecise alignment between high-level semantic concepts and low-level visual features, especially under distribution shifts in open-vocabulary settings.
\ding{184} \textbf{Occlusion-induced Ambiguity}: As depicted in Fig.~\ref{abstract} (d), real-world scenes frequently encounter partial occlusions or suboptimal viewpoints, which truncate critical visual features for interaction verification. Without mechanisms to recover limited visual information, MLLMs may resort to biased textual priors, causing incorrect predictions in ambiguous scenarios.

To address the above-mentioned challenges, we argue that an effective OV-HOI agent requires three synergistic capabilities: 
\ding{182} \textbf{Cognitive Mapping}, which adopts a structured reasoning framework that explicitly models the cognitive mapping between detected entities and possible actions. By bridging the gap from perception to cognition, this structure enables the model to construct comprehensive interaction hypotheses with physical plausibility.
\ding{183} \textbf{Tool Augmenting}. The agent learns to dynamically invoke domain-specific tools for gathering targeted visual-semantic evidence during reasoning. Such an interactive workflow transforms the static foundation model into an interactive agent, effectively reinforcing cross-modal alignment while suppressing hallucinations.
\ding{184} \textbf{Generative World Modeling}, which adopts diffusion models to reconstruct plausible alternative views under occlusion, effectively expanding its observational horizon. This generative prior acts as a world model that empowers the agent to reasonably \textbf{"imagine"} through ambiguity or limited viewpoints. 

Therefore, we introduce ImagineAgent, a novel framework that integrates cognitive reasoning, generative imagination, and \textbf{tool-augmented reinforcement learning (RL)} for robust open-vocabulary HOI detection. 
Specifically, to bridge the perception-to-cognition gap, our method first constructs \textbf{\emph{hicodet-6K} dataset} for \textbf{supervised fine-tuning (SFT)}. \emph{hicodet-6K} is the first chain-of-thought (CoT) dataset in the Open-Vocabulary Human-Object Interaction (OV-HOI) field that simulates both reasoning and cognitive mapping by structuring perceived entities into interaction pairs for comprehensive predictions.
Subsequently, we develop a tool-augmented framework tailored to group relative policy optimization (GRPO) in RL stage. During the inference process, our method dynamically invokes domain-specific tools like online retrieval and image cropping for cross-modal interleaving.

Moreover, to resolve occlusion-induced ambiguity, we further incorporate a generative model as an auxiliary tool to reconstruct alternative viewpoints of the original image, effectively addressing the ambiguities of limitations of its initial perspective.
Furthermore, we propose a composite reward mechanism that jointly optimizes
both the accuracy of HOI predictions and the efficiency of tool utilization.
In summary, our contributions are four-fold:

\begin{itemize}[leftmargin=*,nosep]
    \item \textbf{Tool-Augmented Agentic RL.} ImagineAgent introduces a novel framework that unifies both cognitive mapping reasoning and generative modeling with tool-augmentation reinforcement learning, thereby enabling robust OV-HOI through dynamic, visually grounded reasoning.
    \item \textbf{CoT Cognitive Mapping.} We propose \emph{hicodet-6K} dataset that simulates both reasoning and cognitive mapping by structuring perceived entities into interaction pairs for comprehensive predictions, thus effectively bridging the gap between perception and cognition.
    \item \textbf{Multi-Tool Integration.} We design a multimodal tool library integrating online retrieval, image cropping and generative modeling, which dynamically augments reasoning with domain-specific tools to resolve both cross-modal hallucinations and occlusion-induced image ambiguity. 
    \item \textbf{Empirical Effectiveness.} Experimental results on both SWIG-HOI~\citep{wang2021SWIG-HOI} and HICO-DET~\citep{chao2018HICO-DET} datasets demonstrate that our proposed method achieves state-of-the-art performance in OV-HOI detection, requiring merely approximately 36.7\% of the training data compared to existing methods.
\end{itemize}




\begin{figure*}[t]
  \begin{center}
    \centerline{\includegraphics[width=\textwidth]{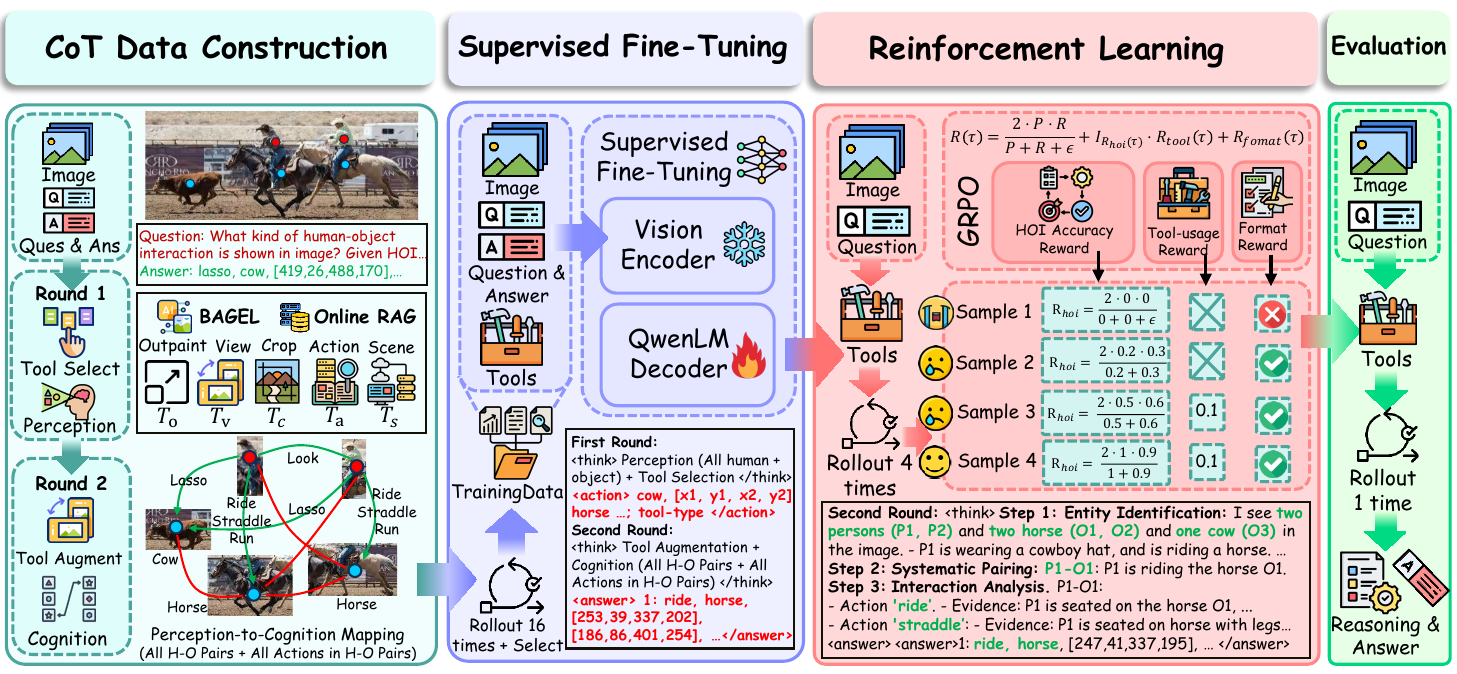}}
    \caption{
    \textbf{Pipeline of our method.} (i) We first construct a high-quality dataset of structured reasoning chains named \emph{hicodet-6K} by performing multiple rollouts with a powerful base model. Each data sample encapsulates a two-round reasoning process: (1) perception and tool selection, and (2) tool augmentation and cognition. (ii) Subsequently, \emph{hicodet-6K} dataset is adopted for SFT to initialize the agent's policy, teaching it the fundamental structure of reasoning. (iii) Following this, the model's policy is further refined through RL using the GRPO algorithm, where the agent learns to make optimal decisions by performing rollouts and receiving feedback from a composite reward that balance HOI predictions, structural coherence and
    tool efficiency, thus achieving robust inference.
    }
    \vspace{-0.3in}
    \label{pipeline}
  \end{center}
\end{figure*}

\section{Methodology}
\textbf{Overview.}
We propose ImagineAgent, a novel framework that integrates cognitive reasoning with generative imagination and tool-augmented RL for robust OV-HOI, as shown in Fig.~\ref{pipeline}. 
Section~\ref{Problem} presents the problem formulation and our workflow. 
Section~\ref{sec:Training} details the \textbf{Training Data Construction}, which synthesizes structured reasoning chain essential for tool invocation and interaction detection.
Section~\ref{sec:SFT} describes the \textbf{Agentic Supervised Fine-tuning}, where the Qwen3-VL base model is pretrained to initialize the agent's policy for tool invocation. 
Section~\ref{sec:RL} outlines the \textbf{Agentic Reinforcement Learning}, which employs the GRPO algorithm to optimize agent policy via a composite reward that balances prediction accuracy and tool efficiency.  

\subsection{Problem Formulation \& Workflow Design}
\label{Problem}
The goal in OV-HOI is to identify and classify comprehensive interactions from a vocabulary unseen during training. The input is a single image $I$ and a query $Q$, while the output is a set of interaction triplets $\mathcal{H} = \{ \langle h_k, o_k, v_k \rangle \}_{k=1}^K$, where $h_k$ is a human, $o_k$ is an object, and $v_k$ is their interaction verb. We deconstruct the problem into a two-stage process:
\begin{itemize}[leftmargin=*,nosep]
\item \textbf{Perception.} This stage involves identifying and localizing all potential human and object instances within the image $I$ to produce sets of localized human instances $\mathcal{I}_H = \{(h_i, b_i)\}_{i=1}^N$ and object instances $\mathcal{I}_O = \{(o_j, b_j)\}_{j=1}^M$, where $b$ denotes the entity's bounding box.
\item \textbf{Cognition}: This stage involves reasoning over the perceived instances to correctly pair entities and identify their interaction. It pairs instances $(h_i, o_j)$ from $\mathcal{I}_H \times \mathcal{I}_O$ and predicts all plausible interaction verb $v_k$, thereby forming each valuable triplet $\langle h_i, o_j, v_k \rangle \in \mathcal{H}$ as final prediction.
\end{itemize} 
To achieve this goal, we design following two-stage workflow to integrates domain-specific tools:

\begin{figure*}[t]
  \begin{center}
    \centerline{\includegraphics[width=\textwidth]{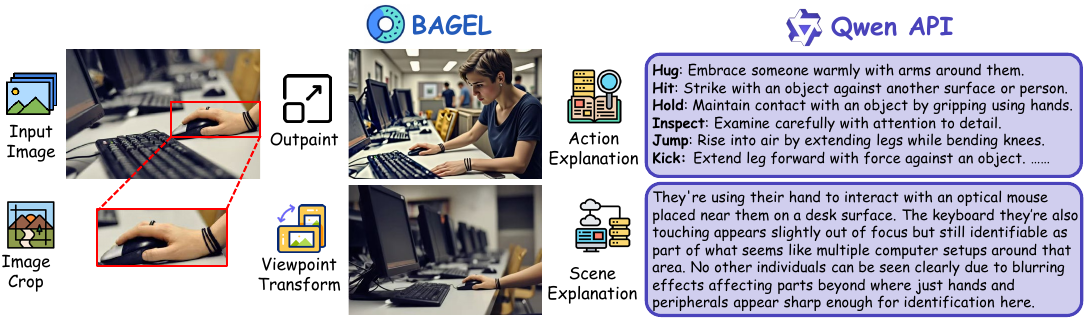}}
    \caption{
    \textbf{Tool Library.} To resolve visual and semantic ambiguities of the image, our framework equips the agent with diverse tools. It leverages the BAGEL model for generative imagination via Outpaint and Viewpoint Transform, the Qwen API for Online RAG through Action Explanation and Scene Explanation, and a standard Image Crop tool to extract focused fine-grained interaction details.
    }
    \vspace{-0.25in}
    \label{tool}
  \end{center}
\end{figure*}

\textbf{Stage 1: Perception \& Tool Selection.}
Given an image $I$ and query $Q$, our model identifies sets of localized human instances $\mathcal{I}_H$ and object instances $\mathcal{I}_O$. Concurrently, it constructs a cognitive map of potential interaction pairs $(h_i, o_j)$. For each potential interaction that exhibits ambiguity or insufficient evidence, our agentic framework formulates a plan by selecting an appropriate toolset from tool library $\mathcal{T}$, which comprises an image cropping tool ($\mathcal{T}_c$), retrieval-augmentation tools for scene and action explanation ($\mathcal{T}_s$ and $\mathcal{T}_a$), and generative tools for viewpoint transformation and outpainting ($\mathcal{T}_v$ and $\mathcal{T}_o$). The policy network $\pi_\theta$ determines the tool call $C \in \mathcal{T}$ based on the initial visual context and the specific interaction hypothesis being evaluated. 

\textbf{Stage 2: Cognition \& Tool Augmentation.}
The second stage executes the selected tool call $C$ and integrates its output to transition from perception to cognition. The tool outputs are not merely concatenated but are adopted to dynamically enhancing agent's scenario understanding. A detailed view from the cropping tool ($\mathcal{T}_c$) allows for focused attention, while contextual knowledge from retrieval tools ($\mathcal{T}_s, \mathcal{T}_a$) enriches the agent's semantic representation of the query. Moreover, an enhanced image $I'$ that resolves prior visual ambiguities from generative tools ($\mathcal{T}_v, \mathcal{T}_o$) provides the agent reasons over an \textit{imagined reality}. The final prediction for an interaction triplet $\langle h, o, v \rangle \in \mathcal{H}$ is then inferred by conditioning the model $\pi_\theta$ on these augmented inputs:
\begin{equation}
    p(\langle h, o, v \rangle | I', Q') = \pi_\theta(I', Q'),
\end{equation}
where $I' = I \oplus I_{\text{aug}}$ and $Q' = Q \oplus E_{\text{aug}}$, $\oplus$ denotes modality-specific augmentation, and $I_{\text{aug}}$ and $E_{\text{aug}}$ respectively represent the augmented visual and textual evidence, while $ \pi_\theta $ maintains the same policy network of MLLMs parameterized by weights $ \theta $.

\textbf{Tool Library.}
As shown in Fig.~\ref{tool}, our framework integrates a diverse tool library to facilitate a multi-stage reasoning process.
\ding{182} \textbf{Image Crop}: to ensure detailed evidence is not overlooked, this tool allows the agent to scrutinize fine-grained interaction details within high-resolution sub-regions. \ding{183} \textbf{Outpaint \& Viewpoint Transform}: to resolve occlusion-induced ambiguity, we equip the agent with generative imagination via BAGEL~\citep{Bagel_2025_arxiv}, allowing it to reconstruct missing context or synthesize alternative views. \ding{184} \textbf{Action \& Scene Explanation}: to mitigate cross-modal hallucination, we leverage the Qwen API for online RAG, which injects precise semantic definitions and contextual information, anchoring predictions in external knowledge. Details are provided in the Appendix~\ref{appx:tool_library}.

\subsection{Training Data Construction}
\label{sec:Training}
A fundamental challenge when applying existing MLLMs to OV-HOI tasks is the absence of explicit reasoning chains. Standard pipelines provide static labels but lack intermediate cognitive steps: \textbf{what} tool to use, \textbf{which} human-object pairs to link, and \textbf{how} plausible verbs are identified. 
To address this, we devise a process to synthesize a high-fidelity structured reasoning chains for constructing our proposed \emph{hicodet-6K} dataset, which creates explicit examples to subtly resolve the above-mentioned cognitive steps, thus effectively bridge the gap between perception and cognition during inference.

\textbf{Data Collection.} Our data generation begins by leveraging the Qwen3-VL-32B~\citep{bai2025qwen3} to explore the HICO-DET training dataset. For each image, we perform 16 independent policy rollouts to generate a diverse set of potential reasoning pathways. 
An instance is considered solvable if at least one rollout produces a triplet set that matches the training annotations under the same matching protocol used for reward computation, including both semantic consistency and spatial IoU constraints. 

\textbf{Data Assessment.} To ensure the quality and reliability of the synthesized data, we introduce a rigorous assessment stage. Specifically, we employ the Qwen-VL-Max model as an expert validator to meticulously review and score each reasoning chain. Chains containing factual errors, logical inconsistencies, or hallucinated reasoning are systematically discarded. This validation procedure yields approximately 6,000 high-fidelity reasoning chains for SFT. The remaining 8,000 trajectories, which were confirmed as successful but not selected in SFT, are then served as the training data for RL stage, providing a wider range of successful behaviors for policy exploration. To prevent data leakage, hicodet-6K and all RL trajectories are constructed only from the HICO-DET training set. 

\textbf{Generating Pipeline.} Our generating pipeline implements a three-step reasoning process. First, based on the initial perception stage, the agent identifies all human instances $\mathcal{I}_H$ and object instances $\mathcal{I}_O$. Second, it constructs a comprehensive hypothesis space by performing an exhaustive pairing of all perceived entities, creating a set of candidate human-object pairs $(h_i, o_j) \in \mathcal{I}_H \times \mathcal{I}_O$. Finally, for each pair, the agent evaluates a constrained set of valid verbs $v_k$, scoring each of them based on visual evidence and contextual cues. Actions with high scores deemed valid and selected to constitute all validated interaction triplets $\mathcal{H} = \{ \langle h_k, o_k, v_k \rangle \}_{k=1}^K$. This structured approach ensures comprehensive consideration of all potential HOIs, effectively bridging perception-to-cognition gap.

\subsection{Agentic Supervised Fine-Tuning}
\label{sec:SFT}
\textbf{After constructing the training data}, we introduce a \textbf{cold-start phase} that prioritizes supervised fine-tuning (SFT) before applying reinforcement learning (RL). 
Inspiring from R1-Zero~\citep{deepseekr1}, we initially attempt direct RL optimization to train our method. However, preliminary experiments reveal a progressive decline in the frequency of tool invocation during policy rollouts. 
This behavior likely arises from a distributional discrepancy in target domain between tool-enhanced visual features and model's pretraining data. Therefore, we opt to perform SFT before RL.

Specifically, We formalize each training instance as $\mathcal{W} = (\mathcal{X}, \mathcal{Y}, \mathcal{Z}, \mathcal{O})$, where $\mathcal{X}$ denotes the input modality, $\mathcal{Y}$ represents task instructions, $\mathcal{Z} = \{z_1,\dots,z_T\}$ captures reasoning steps, and $\mathcal{O}$ is the target output. The objective minimizes the negative log-likelihood of reasoning process:
\begin{equation}
\mathcal{L}_{\text{SFT}} = -\mathbb{E}_{\mathcal{W} \sim \mathcal{D}} \left[ \sum_{t=1}^{T} \log p_{\theta}(z_t \mid \mathcal{X}, \mathcal{Y}, z_{<t}) \right],
\end{equation}
where $p_{\theta}(\cdot)$ denotes conditional probability distribution. 
This formulation enables reasoning chains are aligned with both task instructions and GT during pre-training.

\subsection{Agentic Reinforcement Learning}
\label{sec:RL}
\textbf{Group Relative Policy Optimization (GRPO).}
We adopt the GRPO algorithm~\citep{shao2024deepseekmath} for policy optimization. 
GRPO innovatively employs a groupwise comparison framework to evaluate candidate responses.
Specifically, for each query $ q $ paired with its ground-truth solution $ a $ from dataset $ D $, the algorithm generates a set of rollout trajectories $ \{o_1, o_2, \dots, o_G\} $ based on the previous policy $ \pi_{\theta_{\text{old}}} $. The policy $ \pi_\theta $ is then refined through the optimization of this objective function:  
\begin{equation}
\begin{aligned}
    &\mathcal{L}_{GRPO}(\theta) = -\mathbb{E}_{q \sim P(Q), \{o_i\}_{i=1}^{G} \sim \pi_{\theta_{old}}(O|q)} \Bigg[ \frac{1}{G} \sum_{i=1}^G \\
    &\quad \Bigg( \min\left( \frac{\pi_\theta(o_i|q)}{\pi_{\theta_{old}}(o_i|q)} A_i, \text{clip}\left(\frac{\pi_\theta(o_i|q)}{\pi_{\theta_{old}}(o_i|q)}, 1-\epsilon, 1+\epsilon\right) A_i \right) - \beta \mathbb{D}_{KL}(\pi_\theta \| \pi_{ref}) \Bigg) \Bigg],
\end{aligned}
\end{equation}
\begin{equation}
\mathbb{D}_{KL}(\pi_\theta||\pi_{ref}) = \frac {\pi_{ref}(o_i|q)} {\pi_\theta(o_i|q)} - \log \frac {\pi_{ref}(o_i|q)} {\pi_\theta(o_i|q)} - 1,
\end{equation}
where $ \beta $ is adopted to balance the trade-off between exploration and stability during optimization. Then the advantage estimator $ A_{i} $ is calculated using normalized rewards from the trajectory group:  
\begin{equation}
A_{i} = \frac{r_i - \text{mean}(\{r_1, r_2, \dots, r_G\})}{\text{std}(\{r_1, r_2, \dots, r_G\})}.
\end{equation}
Each trajectory $ o_i $ receives a composite reward through rule-based verification designed as follows.

\textbf{Reward Design.} Effective reward functions should balance precision \& recall of predictions, structural coherence, and tool efficiency. To this end, we design a composite reward that integrates three components: tool-usage reward $R_{\text{tool}}$, format reward $R_{\text{format}}$ and hoi-based accuracy reward $R_{\text{hoi}}$.
The tool-usage reward is defined as $R_{\text{tool}}(\tau)=1$ when at least one valid tool invocation is executed and the generated answer obtains a positive HOI reward, otherwise $R_{\text{tool}}(\tau)=0$. $R_{\text{format}}(\tau)=1$ if the trajectory $\tau$ follows the required structured reasoning format; otherwise $R_{\text{format}}(\tau)=0$. 

Differently, the hoi-based accuracy reward $R_{\text{hoi}}$ quantifies the similarity between the predicted triplets $\mathcal{H}_p = \{\langle h_{p_i}, o_{p_i}, v_{p_i} \rangle\}_{i=1}^{N_p}$ against the ground-truth triplets $\mathcal{H}_g = \{\langle h_{g_i}, o_{g_i}, v_{g_i} \rangle\}_{i=1}^{N_g}$. This requires solving a complex set-to-set matching problem, for which we first compute pairwise affinity score $s_{ij}$ between each predicted triplet $p_i$ and ground-truth triplet $g_j$. 
A non-zero score is assigned only if the pair $(p_i, g_j)$ passes both semantic and spatial verification checks. 
For semantic verification, we encode the verb and object strings using the CLIP ViT-L/14 text encoder and compute cosine similarity in the resulting text embedding space. 
For spatial verification, we use the human and object bounding boxes of the predicted and ground-truth triplets. 
The matching conditions are defined as:
\begin{equation}
\label{eq:match_conditions}
\begin{aligned}
    \text{cos}(v_{p_i}, v_{g_j}) > \delta \quad \land \quad \text{cos}(o_{p_i}, o_{g_j}) > \delta, \\
    \text{IoU}(b_{h_{p_i}}, b_{h_{g_j}}) > \eta \quad \land \quad \text{IoU}(b_{o_{p_i}}, b_{o_{g_j}}) > \eta,
\end{aligned}
\end{equation}
where $\text{cos}(\cdot, \cdot)$ is the cosine similarity,  with $\delta=0.8$ and $\eta=0.5$. The affinity score $s_{ij}$ is defined as 1 if Eqn.~\ref{eq:match_conditions} satisfy, otherwise $s_{ij}=0$. 
With the $N_p \times N_g$ affinity score matrix $S = [s_{ij}]$, we formulate it as a linear assignment problem and employ Hungarian algorithm to find the optimal one-to-one matching. The algorithm operates on a cost matrix $C$, where $c_{ij} = 1 - s_{ij}$, and finds the assignment $M^*$ that minimizes the total cost:
\begin{equation}
M^* = \underset{M \in \Pi(N_p, N_g)}{\arg\min} \sum_{(i,j) \in M} c_{ij},
\end{equation}
where $\Pi(N_p, N_g)$ is the set of all possible one-to-one matchings. From this optimal matching, we identify the number of true positives (TP) as the count of matched pairs with a score greater than zero, i.e., $\text{TP} = |\{(i, j) \in M^* \mid s_{ij} > 0\}|$. The final accuracy reward is then calculated as the F1-score, which harmonizes precision ($P$) and recall ($R$):
\begin{equation}
P = \frac{\text{TP}}{N_p}, \quad R = \frac{\text{TP}}{N_g}, \quad R_{\text{hoi}} = \frac{2 \cdot P \cdot R}{P + R + \epsilon},
\end{equation}
where $\epsilon$ is a small constant to prevent division by zero. Then the final reward is formally defined as:
\begin{equation}
R(\tau) = R_{\text{hoi}} + R_{\text{format}}(\tau) + \mathbb{I}_{R_{\text{hoi}}(\tau) > 0} \cdot R_{\text{tool}}(\tau),
\label{eq:total_reward}
\end{equation}
where $\mathbb{I}_{R_{\text{hoi}}(\tau) > 0}$ is an indicator function. Such composite reward not only maximizes a balanced accuracy metric but also achieves structural coherence and tool-usage efficiency.







\section{Experiment}

\textbf{Datasets}. 
We perform our experiments on two datasets: HICO-DET~\cite{chao2018HICO-DET} and SWIG-HOI~\cite{wang2021SWIG-HOI}. HICO-DET includes 600 interaction types made from 117 human actions and 80 objects. Following previous studies~\cite{hou2020VCL,wang2022_THID}, we exclude 120 rare interactions to test our model in a zero-shot detection setting. SWIG-HOI provides a wide range of interactions with 400 actions and 1,000 objects. This test set naturally contains many unseen combinations. It has about 14K images and 5.5K interaction types, including 1.8K interactions that are not present during training.


\textbf{Evaluation Metrics}. 
Following prior work~\cite{wang2022_THID, lei2024CMD-SE, lei2025_INPCC}, we evaluate our model using mean Average Precision (mAP). A prediction is a true positive if it satisfies: (1) The Intersection over Union (IoU) for both the human and object bounding boxes are greater than 0.5. (2) The predicted verb (interaction) label and corresponding object label are both correct.



\textbf{Implementation Details}. 
We perform experiments with Qwen3-VL-8B as our base model. According to Sec.~\ref{sec:Training}, we respectively select approximately 6,000 and 8,000 training samples for both SFT and RL.
The training was conducted on 8 NVIDIA H20 GPUs (96 GB VRAM) with batch size of 8. The training process took around 40 hours for 1,000 iterations (1 epoch), with 4 rollouts per sample and a learning rate of 5e-7.
More details are in Appendix~\ref{Hyperparameters}.

\subsection{Comparison with SOTA Methods}

\begin{wrapfigure}{r}{0.6\textwidth}
  \centering
  \vspace{-3mm}
  \begin{minipage}[t]{0.6\textwidth}
      \centering
      \captionsetup{type=table}
    \caption{Comparison of our model with existing methods on HICO-DET under the simulated zero-shot setting.}
    \label{tab:hico-det} 
        \resizebox{\textwidth}{!}{
          \begin{tabular}{@{}l|c|ccc@{}}
            \toprule
            Method & {\makecell[c]{Pretrained \\ Detector}} & Unseen & Seen & Full \\
            \midrule
            \multicolumn{1}{@{}c|@{}}{\textit{Zero-shot Methods}} & - & - & - & - \\
            \midrule
            FCL~\cite{hou2021detecting}   & \ding{51}  & 13.16 & 24.23 & 22.01 \\
            GEN-VLKT~\cite{liao2022gen}  &  \ding{51}  & 21.36 & 32.91 & 30.56 \\
            HOICLIP~\cite{ning2023hoiclip}  & \ding{51} & 23.48 & 34.47 & 32.26 \\
            CLIP4HOI~\cite{mao2023clip4hoi} & \ding{51} & \cellcolor{best2}28.47 & \cellcolor{best}{35.48} & \cellcolor{best}{34.08} \\
            HOIGen~\cite{guo2024HOIGen}     & \ding{51} & \cellcolor{best}31.01 & \cellcolor{best2}34.57 & \cellcolor{best2}33.86 \\
            \midrule
            \multicolumn{1}{@{}c|@{}}{\textit{Open-vocabulary Methods}} & - & - & - & - \\
            \midrule
            THID~\cite{wang2022_THID}   & \ding{55}    & 15.53 & 24.32 & 22.38 \\
            CMD-SE~\cite{lei2024CMD-SE} & \ding{55} & 16.70 & 23.95 & 22.35 \\
            INP-CC~\cite{lei2025_INPCC} & \ding{55}  & {17.38} & \cellcolor{best2}{24.74} & \cellcolor{best2}{23.12} \\
            Qwen2.5-VL-7B & \ding{55} & 16.94 & 13.43 & 14.13 \\
            Qwen3-VL-8B (Base) & \ding{55} & \cellcolor{best2}{19.41} & 18.24 & 18.47 \\
            \textbf{ImagineAgent (Ours)} &\ding{55} &  \cellcolor{best}{37.85} &  \cellcolor{best}{36.49} & \cellcolor{best}{36.76} \\
            \bottomrule
          \end{tabular}
          }
    \vspace{-3mm}
  \end{minipage}
\end{wrapfigure}

\textbf{Results on HICO-DET}. 
We evaluate our method on the HICO-DET dataset in Tab.~\ref{tab:hico-det}. Following INP-CC~\cite{lei2025_INPCC}, we categorize existing methods into two groups for fair assessment. 
Note that the \colorbox{best}{best} and \colorbox{best2}{second best} performances are highlighted.  
As can be seen, while zero-shot methods like HOICLIP~\cite{ning2023hoiclip} and CLIP4HOI~\cite{mao2023clip4hoi} have shown strong results, they inevitably rely on detector backbones pre-trained on large-scale object detection datasets like COCO. This pre-training can provide an unfair advantage, as both HICO-DET and COCO datasets share a significant portion of their object categories, limiting the assessment of true generalization in an open-world settings.

Open-vocabulary methods offer a more rigorous test of generalization by forgoing such detection pre-training. ImagineAgent achieves \textbf{36.76\%} mAP on the Full split, a remarkable \textbf{13.64\% mAP} improvement over the prior state-of-the-art, INP-CC~\cite{lei2025_INPCC}. This significant leap demonstrates the powerful reasoning and ambiguity-resolution capabilities of our agentic framework.

\begin{wrapfigure}{r}{0.55\textwidth}
  \centering
  \vspace{-3mm}
  \begin{minipage}[t]{0.55\textwidth}
      \centering
      \captionsetup{type=table}
    \caption{Comparison of our model with existing methods on SWIG-HOI under simulated zero-shot setting.}
    \label{tab:swig-hoi}
        \resizebox{\textwidth}{!}{
          \begin{tabular}{@{}l|cccc@{}}
            \toprule
            Method & Non-rare & Rare & Unseen & Full \\
            \midrule
            QPIC~\cite{tamura2021qpic} & 16.95 & 10.84 & 6.21 & 11.12  \\
            GEN-VLKT~\cite{liao2022gen} & 20.91 & 10.41 & - & 10.87 \\
            MP-HOI~\cite{yang2024MPHOI} & 20.28 & 14.78 & - & 12.61 \\
            THID~\cite{wang2022_THID}  & 17.67  & 12.82 & 10.04 & 13.26 \\
            CMD-SE~\cite{lei2024CMD-SE} &  21.46 & 14.64 & 10.70 & 15.26 \\
            INP-CC~\cite{lei2025_INPCC} & \cellcolor{best2}{22.84} & \cellcolor{best2}{16.74}  & \cellcolor{best2}{11.02} & \cellcolor{best2}{16.74} \\  
            Qwen2.5-VL-7B & 1.55 & 3.54 & 5.82 & 3.61 \\
            Qwen3-VL-8B (Base) & 3.09 & 7.05 & 11.61 & 7.25 \\          
            \textbf{ImagineAgent (Ours)} & \cellcolor{best}{26.88} & \cellcolor{best}{21.16} & \cellcolor{best}{14.70} & \cellcolor{best}{20.84}  \\
            \bottomrule
          \end{tabular}
          }
    \vspace{-3mm}
  \end{minipage}
\end{wrapfigure}

\textbf{Results on SWIG-HOI}. 
We further evaluate our method on the SWIG-HOI dataset. As shown in Tab.~\ref{tab:swig-hoi}, our method consistently surpasses all previous methods, achieving a Full mAP of \textbf{20.84\%}, outperforming INP-CC by \textbf{4.10\%} mAP. The strength of our approach is further evidenced by its performance on novel and infrequent interactions, with a \textbf{14.70\%} mAP on the Unseen split and \textbf{21.16\%} on the Rare split. Such results validate our robustness, which excels at detecting interactions in open-vocabulary long-tail scenes.

\begin{wrapfigure}{r}{0.6\textwidth}
  \centering
  \vspace{-3mm}
  \begin{minipage}[t]{0.6\textwidth}
      \centering
      \captionsetup{type=table}
    \caption{Ablation of training stages on HICO-DET dataset.}
    \label{tab:ablation_stage}
        \resizebox{\textwidth}{!}{
          \begin{tabular}{@{}l|c|ccc@{}}
            \toprule
            Method & Training Steps & Unseen & Seen & Full \\
            \midrule 
            \emph{(a)} Qwen3-VL-8B & 0 steps & 19.41 & 18.24 & 18.47 \\
            \emph{(b)} w/ SFT (\emph{hicodet-6K}) & 750 steps & \cellcolor{best2}28.24 & \cellcolor{best2}26.98 & \cellcolor{best2}27.23 \\
            \emph{(c)} w/ RL (GRPO) & 1,000 steps & 25.33 & 24.34 & 24.54 \\
            \midrule
            {\textbf{ImagineAgent (Ours)}} & {1,750 steps} & \cellcolor{best}{37.85} &  \cellcolor{best}{36.49} & \cellcolor{best}{36.76} \\
            \bottomrule
          \end{tabular}
          }
  \end{minipage}
\end{wrapfigure}

\subsection{Ablation Study}

\textbf{Effect of Training Stages.} We analyze each training stage's contribution in Tab.~\ref{tab:ablation_stage}. The baseline model (row a) struggles with the HOI task. SFT alone (row b) significantly boosts performance to \textbf{27.23\%} mAP by teaching model structured reasoning. Conversely, RL-only training (row c) shows limited improvement, highlighting the challenge of exploring the vast policy space without a strong prior. Differently, combining SFT and RL (row d) achieves the highest performance at \textbf{36.76\%} mAP. This confirms that SFT provides a crucial cognitive foundation for subsequent RL refinement.

\begin{wrapfigure}{r}{0.5\textwidth}
  \centering
  \vspace{-3mm}
  \begin{minipage}[t]{0.5\textwidth}
      \centering
      \captionsetup{type=table}
    \caption{Ablation of tool-usage on HICO-DET.}
    \label{tab:ablation_tool}
        \resizebox{\textwidth}{!}{
          \begin{tabular}{@{}l|ccc@{}}
            \toprule
            Method  & Unseen & Seen & Full \\
            \midrule
            \emph{(a)} w/o image crop & 37.69 & 35.19 & 35.69 \\
            \emph{(b)} w/o action explanation & 37.37 & 34.17 & 34.81 \\
            \emph{(c)} w/o scene explanation & 37.31 & \cellcolor{best2}{35.96} & \cellcolor{best2}{36.23} \\
            \emph{(d)} w/o view transform &  \cellcolor{best2}{37.82} & 34.90 & 35.48 \\
            \emph{(e)} w/o outpaint & 37.58 & 35.10 & 35.61 \\
            \midrule
            {\textbf{ImagineAgent (Ours)}} & \cellcolor{best}{37.85} & \cellcolor{best}{36.49} & \cellcolor{best}{36.76} \\
            \bottomrule
          \end{tabular}
          }
  \end{minipage}
\end{wrapfigure}

\textbf{Effect of Tool-Usage.} We investigate each tool's contribution in Tab.~\ref{tab:ablation_tool}, where disabling any single tool causes performance degradation. Removing generative tools like viewpoint transformation (d) and outpainting (e) hinders reasoning under occlusion, causing drops to \textbf{35.48\%} and \textbf{35.61\%} mAP. Excluding retrieval-based explanations (b, c) impairs semantic disambiguation. Disabling image cropping (a) leads to a significant drop to \textbf{35.69\%} mAP, indicating that fine-grained analysis is indispensable. The full framework's superior performance validates our multi-tool integration strategy.

\begin{wrapfigure}{r}{0.5\textwidth}
  \centering
  \vspace{-3mm}
  \begin{minipage}[t]{0.5\textwidth}
      \centering
      \captionsetup{type=table}
    \caption{Ablation of reward func. on HICO-DET.}
    \label{tab:ablation_reward}
        \resizebox{\textwidth}{!}{
          \begin{tabular}{@{}l|ccc@{}}
            \toprule
            Method  & Unseen & Seen & Full \\
            \midrule
            \emph{(a)} w/o HOI Reward      & 33.75 & 32.34 & 32.62 \\
            \emph{(b)} w/o Precision \& Recall  & 34.56 & 33.42 & 33.65 \\
            \emph{(c)} w/o Hungarian       & 35.62 & 34.53 & 34.75 \\
            \emph{(d)} w/o Verb Similarity      & \cellcolor{best2}{37.25} & \cellcolor{best2}{36.03} & \cellcolor{best2}{36.27} \\
            \emph{(e)} w/o Tool Reward     & 36.62 & 35.42 & 35.66 \\
            \midrule
            {\textbf{ImagineAgent (Ours)}} & \cellcolor{best}{37.85} & \cellcolor{best}{36.49} & \cellcolor{best}{36.76} \\
            \bottomrule
          \end{tabular}
          }
  \end{minipage}
\end{wrapfigure}

\textbf{Effect of Reward Functions.} Our composite reward's design is pivotal for guiding the agent's policy. As shown in Tab.~\ref{tab:ablation_reward}, removing the core HOI accuracy reward (a) causes a catastrophic drop to \textbf{32.62\%} mAP. Simplifying the accuracy metric by removing the Hungarian algorithm (c) or verb similarity check (d) also degrades performance, confirming that our nuanced F1-score is essential for evaluating complex scenes. Finally, excluding the tool-usage reward (e) drops mAP to \textbf{35.66\%}, showing its importance in encouraging efficient tool invocation. Each component is thus crucial for steering the agent towards accurate and efficient behavior.

\begin{wrapfigure}{r}{0.5\textwidth}
  \centering
  \vspace{-3mm}
  \begin{minipage}[t]{0.5\textwidth}
      \centering
      \captionsetup{type=table}
    \caption{Effect of different generative models on HICO-DET.}
    \label{tab:ablation_edit_model}
        \resizebox{\textwidth}{!}{
          \begin{tabular}{@{}l|ccc@{}}
            \toprule
            Editing Model  & Unseen & Seen & Full \\
            \midrule
            \emph{(a)} FLUX.1 Kontext~\cite{FLUX_Kontext_arxiv2025} & \cellcolor{best}{38.34} & \cellcolor{best}{36.95} & \cellcolor{best}{37.23}  \\
            \emph{(b)} Qwen-Image~\cite{qwen_image_arxiv2025} & \cellcolor{best2}{38.19} & \cellcolor{best2}{36.83} & \cellcolor{best2}{37.10}  \\
            \emph{(c)} BAGEL~\cite{Bagel_2025_arxiv} (\textbf{Ours}) & {37.85} & {36.49} & {36.76}  \\
            \bottomrule
          \end{tabular}
          }
  \end{minipage}
\end{wrapfigure}

\textbf{Effect of Generative Models.} 
We analyze the impact of applying different generative models in Tab.~\ref{tab:ablation_edit_model}. Our investigation reveals that while more advanced models such as FLUX.1 Kontext (a) and Qwen-Image (b) can yield marginal performance improvements, the former reaching \textbf{37.23\%} mAP while come at a greater computational cost. Differently, BAGEL (c), provides a compelling balance between performance and processing speed, makes it a more practical choice for our workflow.

\begin{figure*}[t]
  \begin{center}
    \centerline{\includegraphics[width=\textwidth]{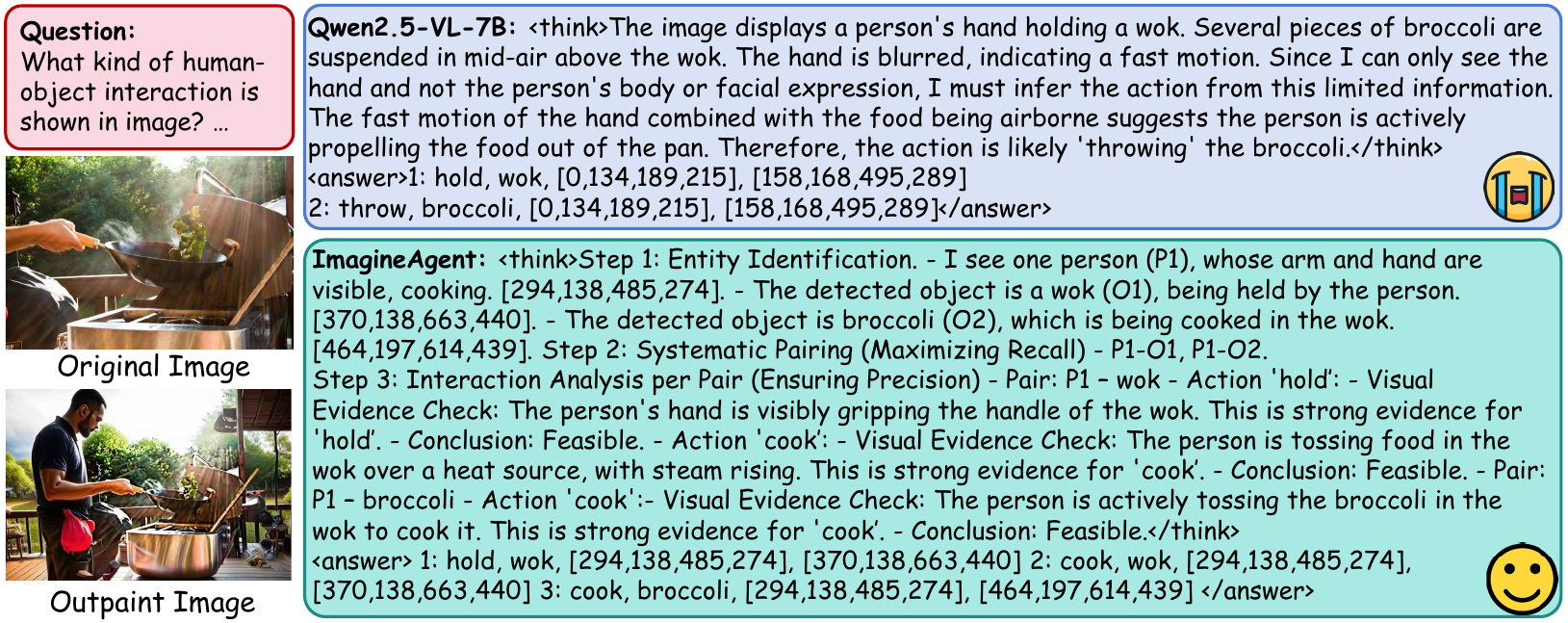}}
    \caption{
    \textbf{Case Study between Qwen2.5-VL-7B and our method.} Our method can accurately identify comprehensive HOI predictions, while Qwen2.5-VL-7B misclassifies the interaction labels.
    }
    \vspace{-0.25in}
    \label{case1}
  \end{center}
\end{figure*}

\subsection{Qualitative Results}
Our method utilizes outpainting, view transformation, and cropping to respectively enrich context, resolve occlusion, and focus on critical details (see Appendix~\ref{appx:case_study} for more cases). This enhanced perception directly translates to superior cognitive reasoning, as shown in Fig.~\ref{case1}, where our method accurately deduces a comprehensive set of correct interactions. To mitigate potential hallucinations from generative images, we append designed prompt instructions to govern the use of synthesized images (see Appendix~\ref{Generation} for details). 



\section{Related Work}
\textbf{Open-Vocabulary HOI Detection.}
Previous methods for open-vocabulary Human-Object Interaction (HOI) detection primarily leverage the cross-modal alignment capability of CLIP~\cite{liao2022gen, ning2023hoiclip, mao2023clip4hoi}, and utilize DETR~\cite{carion2020end} as a critical component to perform object detection in the first place~\cite{kim2025locality}. With the advances of Large Language Models (LLMs), recent literature~\cite {lei2024ez, lei2024CMD-SE, lei2025_INPCC} proposed to improve the fine-grained understanding of diverse HOI concepts through prompting LLMs. Furthermore, MLLMs~\cite{bai2025qwen2, bai2025qwen3} show great potential in open-vocabulary HOI detection in an end-to-end manner with strong visual grounding ability and world knowledge. Thus, we build \textbf{ImagineAgent} upon MLLMs and further achieve tool-augmented RL for robust open-vocabulary inference.



\textbf{Multimodal LLMs Reasoning.}
Recent advances in Large Language Models (LLMs) have demonstrated that RL-based post-training can significantly enhance reasoning capabilities, as exemplified by OpenAI-o1~\citep{jaech2024openaio1} and DeepSeek-R1~\citep{deepseekr1}. These paradigms have been extended to Multimodal Large Language Models (MLLMs) for tasks like mathematical VQA~\citep{peng2025lmmr1}, image segmentation~\citep{liu2025seg}, and video understanding~\citep{feng2025videor1}. However, existing methods struggle with long-sequence hallucination~\citep{chen2025longvilar1} and limited cross-modal interaction. To address this, we propose multimodal CoT reasoning via tool-augmented RL to explicitly reduce cross-modal hallucination.

\textbf{Tool-Augmented Agentic System.}
Recent advancements in MLLMs have shown that external tools can enhance multimodal reasoning. Early works like FAST~\citep{sun2025fast} and MVoT~\citep{li2025MVoT} introduce visual evidence into reasoning, forming multimodal CoT for image tasks. LLaVA-Plus~\citep{liu2023llavapluslearningusetools} pioneered training strategies for tool use, while VPD~\citep{hu2024visual} leveraged program-derived data to transfer tool skills. Recent methods like TACO~\citep{ma2024taco} and PyVision~\citep{zhao2025pyvision} expand tool use with RL. However, existing approaches rely on static tool pipelines, limiting their ability to complex, open-vocabulary settings. Our framework addresses this by unifying both cognitive mapping reasoning and generative modeling with tool-augmented RL for robust OV-HOI prediction.


\section{Conclusion}
In this work, we introduced \textbf{ImagineAgent}, a novel agentic framework designed to enhance OV-HOI understanding through structured reasoning and multimodal augmentation.
Our approach leverages four key components: (i) The \emph{hicodet-6K} CoT dataset for SFT, bridging the perception-to-cognition gap by structuring perceived entities into interaction pairs; (ii) A multimodal tool library integrating online retrieval, image cropping, and generative modeling, enabling dynamic reasoning with domain-specific tools to resolve ambiguities and hallucinations; (iii) A generative model that reconstructs alternative viewpoints, allowing the agent to “imagine” beyond limited perspectives; and (iv) A composite reward mechanism that jointly optimizes prediction accuracy and tool efficiency. Extensive evaluations on the SWIG-HOI and HICO-DET datasets validate our SOTA performance while using merely 36.7\% of the training data compared to existing methods. Future work will explore expanding the tool library and applying imagination-driven reasoning to broader complex visual tasks.


\bibliographystyle{plainnat}
\bibliography{references}

\newpage

\appendix
\section*{Appendix}

\label{app:discussion}
In this appendix, we provide more experimental details, tool library details, and prompt details of our proposed method.
Specific detailed contents are as follows:

\setlength{\cftbeforesecskip}{0.5em}
\cftsetindents{section}{0em}{1.8em}
\cftsetindents{subsection}{1em}{2.5em}
\etoctoccontentsline{part}{Appendix}
{
  \etocsettocstyle{}{}
  \localtableofcontents
}

\section{Experiment}
\subsection{Full Experimental Setup (Reproducibility)}
\subsubsection{Datasets and evaluation protocols}


\textbf{HICO-DET}
We conduct experiments on HICO-DET, a standard benchmark for HOI detection with 600 interaction categories composed of 117 verbs and 80 object classes and use the official test set for evaluation. Following prior zero-shot HOI protocols (CMD-SE~\citep{lei2024CMD-SE} and INP-CC~\cite{lei2025_INPCC}), 120 rare interaction categories are excluded from training to simulate the zero-shot setting, where these categories are treated as Unseen, and the remaining categories are treated as Seen. Results on Unseen, Seen, and Full splits are reported.

\textbf{SWIG-HOI}
We also evaluate on SWIG-HOI, a large-vocabulary HOI benchmark built upon the publicly available SWiG dataset to study large-vocabulary and open-vocabulary HOI detection. SWIG-HOI contains more than 400 action (verb) categories and 1,000 object categories, substantially expanding the label space compared to earlier datasets such as HICO-DET. In our setting, Unseen refers to out-of-vocabulary action types that do not appear in the training set, and we report results on the standard Non-rare, Rare, Unseen, and Full splits.

\textbf{Metric.}
We follow prior work and evaluate using mean Average Precision (mAP) computed by the official evaluation scripts, with no additional filtering.

\textbf{True positive criterion.}
A predicted HOI triplet is considered correct if it satisfies both:
(i) IoU > 0.5 for the predicted human bounding box and its matched ground-truth human box, and IoU > 0.5 for the predicted object bounding box and its matched ground-truth object box.
(ii) the predicted interaction label matches the ground truth.

\subsubsection{Implementation Details and Hyperparameters}
\label{Hyperparameters}

\textbf{Base Model.}
We choose Qwen3-VL-8B as our backbone and fine-tune the model using full-parameter finetuning.

\textbf{Input Configuration.}
For each sample, the maximum image resolution is set to 401408 pixels,
and the number of image tokens is capped at 512.
The maximum text sequence length is 8192 tokens, and the maximum generation length during rollouts is 4096 tokens.
We use resize and center crop for image preprocessing.

\textbf{Optimization.}
We train the model with Adam optimizer using a learning rate of 1e-6 for SFT Stage, and 5e-7 for RL Stage with weight decay 0.01, and gradient clipping with max norm 5.
All experiments are conducted with BF16 mixed precision, and the random seed is fixed to 1234 for reproducibility.

\textbf{Batching and Training Budget.}
We use a per-device batch size of 1 on 8 GPUs,
with gradient accumulation steps 1 for both SFT and RL.
The resulting global batch size is therefore
$\textbf{8} = \textbf{1} \times \textbf{8} \times \textbf{1}$ for SFT, and $\textbf{8} = \textbf{1} \times \textbf{8} \times \textbf{1}$ for RL.

\textbf{SFT Stage.}
In supervised fine-tuning (SFT), we train for 1 epoch and use 5999 trajectories.

\textbf{RL Stage (GRPO).}
In agentic reinforcement learning, we optimize the policy using GRPO for approximately 1,000 steps.
For each training query, we sample 4 rollouts.
We use KL regularization coefficient $\beta = 0.04$.
The reference policy $\pi_{\mathrm{ref}}$ is chosen as SFT model, and the old policy $\pi_{\theta_{\mathrm{old}}}$ is updated every 1 steps.

\textbf{Hardware.}
Training is performed on eight \textbf{Nvidia H20} GPUs with \textbf{96GB} memory for each GPU.
The total training time is approximately \textbf{2.5 hours} for SFT and \textbf{40 hours} for RL.

%

\subsection{More Ablation Studies}
\label{sec:more_ablation}

\textbf{Comparison with Frontier MLLMs.}
To further evaluate whether general-purpose frontier MLLMs can directly solve fine-grained OV-HOI reasoning, we compare ImagineAgent with several strong proprietary models, including GPT-4o, Gemini 3.0 Pro, and GPT-5. As shown in Tab.~\ref{tab:frontier_mllm}, although these frontier models possess strong native multimodal reasoning abilities, their performance remains limited on fine-grained HOI comprehension. In particular, GPT-5 achieves only 23.18\% Full mAP, which is still substantially lower than ImagineAgent with Qwen3-VL-8B as the backbone.

This result suggests that parameter scaling alone is insufficient for resolving the severe occlusions, spatial ambiguities, and human-object pairing challenges in OV-HOI. In contrast, our specialized tool-augmented reasoning framework explicitly decomposes the task into perception, tool selection, and interaction-aware cognition, thereby providing more reliable visual grounding and structured reasoning. These results further validate the necessity of specialized agentic reasoning for HOI tasks.

\vspace{-0.1in}
\begin{table}[ht]
\centering
\caption{Comparison with frontier MLLMs on HICO-DET under the simulated zero-shot setting.}
\vspace{0.1in}
\label{tab:frontier_mllm}
\resizebox{0.5\textwidth}{!}{
\begin{tabular}{l|ccc}
\toprule
Method & Unseen & Seen & Full \\
\midrule
GPT-4o & 22.15 & 21.45 & 21.78 \\
Gemini 3.0 Pro & 22.45 & 21.75 & 22.06 \\
GPT-5 & \cellcolor{best2}23.45 & \cellcolor{best2}22.85 & \cellcolor{best2}23.18 \\
\textbf{ImagineAgent (Ours)} & \cellcolor{best}{33.55} & \cellcolor{best}{32.40} & \cellcolor{best}{32.84} \\
\bottomrule
\end{tabular}
}
\end{table}

\textbf{Prevalence of Hallucination and Visual Ambiguity.}
We further analyze how often the challenges illustrated in Fig.~\ref{abstract} occur in real OV-HOI evaluation scenarios. Specifically, we evaluate Qwen-VL-Max on the full HICO-DET test set containing 9,546 images and regard a prediction as hallucinated if the model fails to cover all ground-truth action and object categories in its output set. In addition, we randomly sample 600 images from HICO-DET and manually categorize whether each image contains significant object occlusion or severe viewpoint limitation.

As shown in Tab.~\ref{tab:ambiguity_statistics}, hallucination occurs in 38.7\% of the full test set, indicating that native MLLM reasoning frequently fails to fully ground HOI predictions in visual evidence. Meanwhile, 12.3\% of the sampled images contain significant object occlusion, and 19.7\% suffer from severe viewpoint limitations. These observations confirm that both cross-modal hallucination and visual ambiguity are prevalent in OV-HOI, motivating the need for tool-augmented reasoning and generative imagination.

\vspace{-0.1in}
\begin{table}[ht]
\centering
\caption{Statistics of hallucination and visual ambiguity on HICO-DET.}
\vspace{0.1in}
\label{tab:ambiguity_statistics}
\resizebox{0.5\textwidth}{!}{
\begin{tabular}{l|c|c}
\toprule
Metric & Num. Images & Rate \\
\midrule
MLLM Hallucination & \cellcolor{best}9,546 & \cellcolor{best}38.7\% \\
Object Occlusion & 600 & 12.3\% \\
Viewpoint Limitation & \cellcolor{best2}600 & \cellcolor{best2}19.7\% \\
\bottomrule
\end{tabular}
}
\end{table}

\textbf{Error Analysis.}
To better understand the remaining failure cases of ImagineAgent, we randomly sample 600 examples from HICO-DET and manually categorize the major error types. As reported in Tab.~\ref{tab:error_analysis}, incomplete HOI matching accounts for the largest proportion of errors, reaching 39.7\%. This indicates that the model still struggles to exhaustively identify all valid human-object interaction pairs, especially in crowded or visually complex scenes. Object classification and action classification errors account for 25.5\% and 23.2\%, respectively, suggesting that fine-grained semantic discrimination remains challenging. Grounding errors account for 20.8\%, reflecting the difficulty of accurately localizing interacting humans and objects under occlusion or small-object scenarios. These findings reveal that although ImagineAgent significantly improves OV-HOI reasoning, its remaining errors are mainly caused by incomplete interaction coverage and fine-grained semantic confusion. 

\vspace{-0.1in}
\begin{table}[ht]
\centering
\caption{Error analysis on 600 randomly sampled HICO-DET examples.}
\vspace{0.1in}
\label{tab:error_analysis}
\resizebox{0.42\textwidth}{!}{
\begin{tabular}{l|c}
\toprule
Error Type & Proportion \\
\midrule
Action classification error & 23.2\% \\
Grounding error & 20.8\% \\
Object classification error & \cellcolor{best2}25.5\% \\
Incomplete HOI matching & \cellcolor{best}39.7\% \\
\bottomrule
\end{tabular}
}
\end{table}




%

\subsection{Data Construction Details (Reasoning Chains)}
\label{sec:supp_data_construction}

This section details the reproducible pipeline used to synthesize structured reasoning chains for HOI comprehension on HICO-DET. Our construction is implemented as a \textbf{two-turn agentic prompting} process with optional auxiliary images and optional textual knowledge, followed by trajectory packaging into training-ready JSON.

\subsubsection{Generation Prompts}
\label{sec:supp_gen_prompts}

\textbf{Prompt design (two-turn decomposition).}
We generate reasoning chains using a two-turn chat template with \textbf{Qwen3-VL-32B-Instruct} (served by vLLM). Turn-1 performs \emph{object set selection} and \emph{tool selection}, and Turn-2 performs \emph{exhaustive HOI enumeration} under strict constraints and outputs HOI triplets with bounding boxes.

\textbf{Turn-1 (Perception \& Tool Decision).}
Given the input image, the model is instructed to:
(i) detect all visible objects from the predefined HICO object list.
(ii) decide which tools to use among \{\textit{image\_description}, \textit{action\_description}, \textit{outpaint}, \textit{viewpoint\_transform}, \textit{image\_crop}\}.
The output format is enforced as:
\textit{object\_1, [ox1,oy1,ox2,oy2], object\_2, [ox1,oy1,ox2,oy2], ... ; tool\_1, tool\_2, ...}
wrapped by \textit{<think>} and \textit{<answer>} tags.
For training-data synthesis only, we additionally provide \emph{ground-truth objects extracted from the HOI annotation string} inside the prompt to bias the chain toward GT-aligned decisions (used for CoT supervision rather than evaluation).

\textbf{Turn-2 (Constrained HOI Chain-of-Thought).}
Turn-2 takes the original image (always) plus up to two optional auxiliary images (outpaint and/or viewpoint transform) depending on Turn-1 tool decisions. It also injects:
(i) \textbf{DETECTED OBJECTS} from Turn-1.
(ii) a per-object \textbf{VALID ACTIONS} constraint list derived from HICO interactions (i.e., allowed verbs for each object class).
(iii) optional textual evidence, including an image-level description and/or action definitions if the corresponding tools were selected.

Inside \textit{<think>}, the model must follow a fixed structure:
(1) entity listing with person/object boxes,
(2) exhaustive person--object pairing,
(3) per-pair verb feasibility checks restricted to valid verbs,
(4) final review and compilation.
Inside \textit{<answer>}, the model must emit HOIs as a comma-separated list of records:
\textit{idx: action, object, [px1,py1,px2,py2], [ox1,oy1,ox2,oy2]}.
For synthesis, we include the ground-truth HOI string in the prompt as a reference target (the model is instructed to build a reasoning bridge that leads to the GT).

\textbf{Tool library and how it is realized in our code.}
Unlike an online tool-execution agent, our synthesis pipeline implements tool augmentation \emph{by conditional evidence injection}:
\begin{itemize}[leftmargin=*,nosep]
    \item \textbf{\textit{image\_description}}: if selected in Turn-1, we load a precomputed image-level analysis from \textit{train\_all\_description.json} and append it into the Turn-2 prompt.
    \item \textbf{\textit{action\_description}}: if selected, we load per-action textual definitions from \textit{hico\_actions\_rephase\_20.json}. To avoid leaking irrelevant knowledge, we only include definitions for actions appearing in the valid-action constraint set of the current sample.
    \item \textbf{\textit{outpaint}} / \textbf{\textit{viewpoint\_transform}}: We explicitly instruct the model that auxiliary images are \emph{only} for disambiguating actions, and that all bounding box coordinates must be taken from the original image.
    \item \textbf{\textit{image\_crop}}:  if selected, we extract high-resolution sub-regions centered on specific human-object pairs from the original image and inject them as auxiliary visual evidence into the Turn-2 prompt. 
\end{itemize}

\textbf{Independent rollouts and decoding.}
In the released synthesis code, we generate \textbf{one} trajectory per image-query pair with stochastic decoding:
temperature $0.8$, and max generation length $4096$ tokens for both turns.

\textbf{Randomness control.}
The synthesis code relies on the underlying sampler randomness from vLLM at the process level and does not explicitly set per-sample random seeds. All other decoding hyperparameters are kept fixed across samples to ensure reproducibility given the same runtime environment and model weights.

\subsubsection{Expert Validation and Filtering}
\label{sec:supp_expert_validation}

\textbf{Trajectory packaging (what is saved).}
For each HICO training instance, we store:
\textit{detected\_objects}, \textit{selected\_tools}, \textit{first\_turn\_output}, and \textit{second\_turn\_output}.
We also store the exact textual prompt strings for both turns to enable prompt-level reproduction.

\textbf{Expert validation.}
In our overall framework, we employ \textbf{Qwen-VL-Max} as an expert validator to score and filter synthesized chains. Concretely, the validator checks whether the generated chain is (i) visually grounded, (ii) logically consistent with the constrained action set and the claimed evidence, and (iii) format-compliant with the required \textit{<think>/<answer>} schema.
Chains that fail validation (e.g., hallucinated entities/interactions, invalid format, or ungrounded claims) are removed before forming the final SFT/RL corpora.

\textbf{Hard constraints used in construction (rule-based).}
Even before expert scoring, we apply strict rule constraints embedded in the prompts and enforced by parsing:
\begin{itemize}
    \item \textbf{Closed-set object constraint:} all objects must come from the HICO object vocabulary.
    \item \textbf{Per-object verb constraint:} verbs must come from the HICO interaction-derived mapping \textit{object\_to\_verbs}.
    \item \textbf{Bounding box constraint:} all coordinates must correspond to the \emph{original image}, and auxiliary images are prohibited from providing coordinates.
    \item \textbf{Format constraint:} outputs must contain parsable \textit{<answer>} blocks, otherwise the sample is marked invalid.
\end{itemize}



\subsubsection{Examples of Synthesized Reasoning Chains}
\label{sec:supp_examples}

Each synthesized sample contains: (i) the original image (and optional auxiliary images), (ii) Turn-1 object/tool decisions, (iii) Turn-2 structured reasoning in \textit{<think>} and HOI triplets in \textit{<answer>}.

\subsection{More Case Studies}
\label{appx:case_study}
\begin{figure*}[ht]
  \begin{center}
    \centerline{\includegraphics[width=\textwidth]{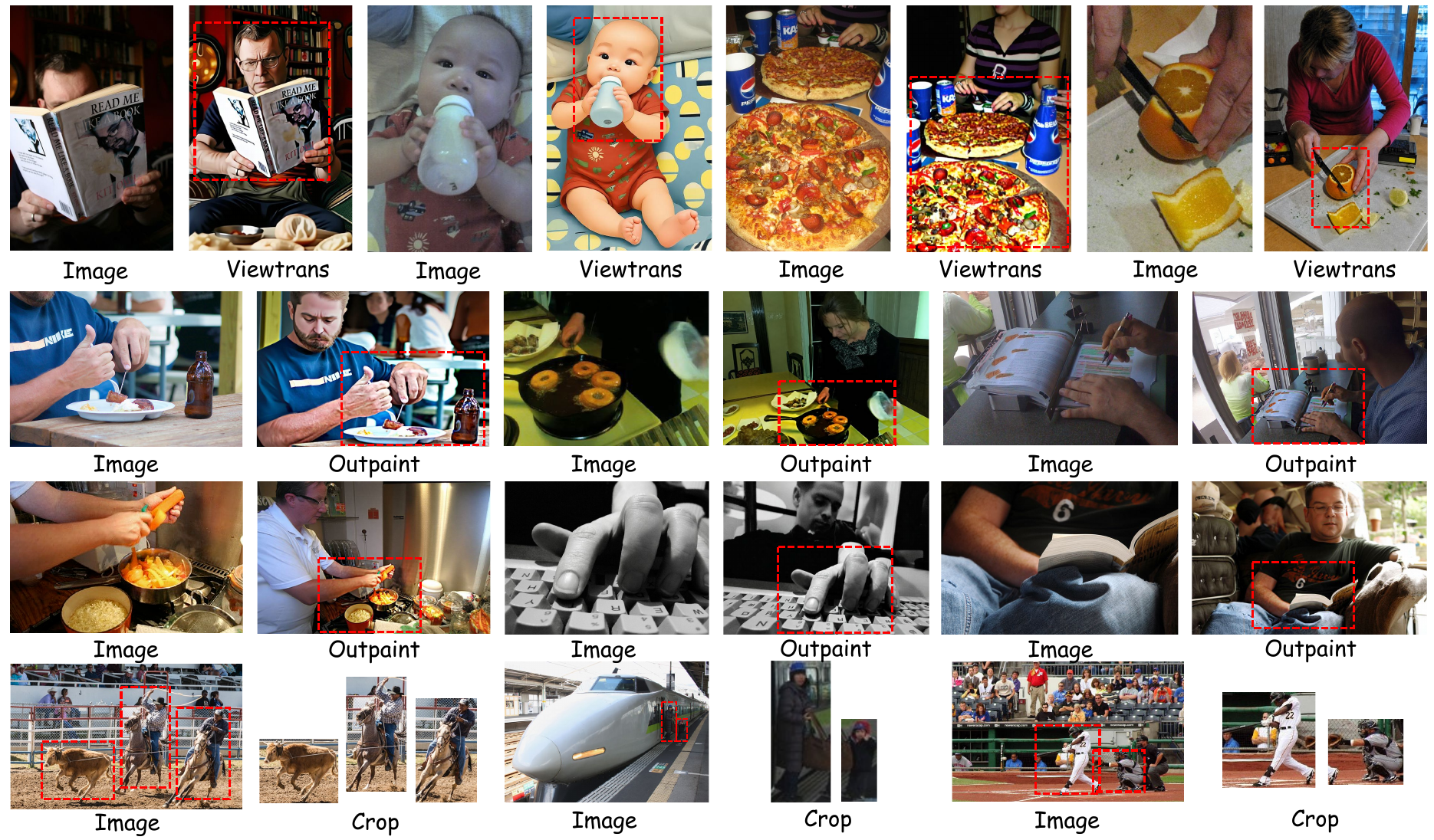}}
    \caption{
    \textbf{Qualitative results of ImagineAgent's generative imagination.} Respectively, \textbf{Image Outpaint} is used to extend the scene's context, providing a more holistic view of the environment that aids in understanding the overall activity. To overcome severe occlusion, the agent employs \textbf{Image View Transformation} to synthesize a novel perspective of the interaction. Furthermore, \textbf{Image Crop} is utilized for a focused analysis of fine-grained interactions, scrutinizing critical details for detection. 
    }
    \vspace{-0.2in}
    \label{result2}
  \end{center}
\end{figure*}

As shown in Fig.~\ref{result2}, our method utilizes outpainting, view transformation, and cropping to respectively enrich context, resolve occlusion, and focus on critical details.
As shown in Fig.~\ref{fig:case3} and Fig.~\ref{fig:case2}, we present more case studies between Qwen2.5-VL-7B and our model on the HICO-DET dataset. These examples highlight the baseline model's susceptibility to cross-modal hallucination, where it generates plausible but incorrect interactions. In contrast, ImagineAgent employs a structured cognitive process, systematically evaluating entity pairs against visual evidence enriched by tools like Outpaint. This allows it to discard flawed hypotheses and accurately identify a comprehensive set of correct interactions, demonstrating superior grounded reasoning.

\section{Tool Library Details}
\label{appx:tool_library}
This section provides a detailed specification of the multimodal tools integrated into our ImagineAgent framework. These tools are designed to solve ambiguities encountered during HOI prediction and are categorized into visual refinement, generative imagination, and online retrieval-augmentation tools.

\vspace{-0.1in}
\begin{table}[ht]
\centering
\caption{Implementation details of the tool library.}
\vspace{0.1in}
\label{tab:tool_details}
\resizebox{\textwidth}{!}{
\begin{tabular}{cccc}
\toprule
Tool & Input & Output & Failure Handling \\
\midrule
Image Crop & predicted human/object boxes & cropped image patch & ignored if boxes are invalid \\
Action Explanation & candidate action phrase & textual action definition & fallback to original query \\
Scene Explanation & image and scene query & textual scene context & fallback to original query \\
Viewpoint Transform & image and editing prompt & transformed image & used only as auxiliary evidence \\
Outpaint & image and editing prompt & expanded image & used only as auxiliary evidence \\
\bottomrule
\end{tabular}}
\end{table}

\subsection{Visual Refinement Tools}
To ensure that fine-grained interaction details are not overlooked in complex scenes, our framework incorporates an \textbf{Image Crop} tool ($T_c$). Its primary function is to allow the agent to isolate and scrutinize high-resolution sub-regions of the input image. This tool is dynamically invoked when the initial perception suggests a potential interaction, but the visual evidence is too small or subtle to be conclusively identified from a global view. By extracting a focused patch centered on a candidate human-object pair, the agent can perform a more detailed analysis of subtle cues, such as hand-object contact or the specific orientation of a tool. This capability is critical for accurately distinguishing between similar interactions and directly mitigates failures arising from the perceptual-cognitive gap, where global context is insufficient for precise grounding.

\subsection{Online Retrieval-Augmentation Tools}
To mitigate cross-modal hallucinations, we leverage the \textbf{Qwen API} as an online Retrieval-Augmented Generation (RAG) tool. This component anchors the agent's reasoning in external, factual knowledge, preventing over-reliance on learned textual priors. It serves two key functions: \textbf{Action Explanation} ($T_a$) and \textbf{Scene Explanation} ($T_s$). When faced with semantically similar verbs, the agent can invoke Action Explanation to retrieve precise definitions and distinguishing characteristics, thus enabling a more nuanced classification. Concurrently, the Scene Explanation function provides high-level contextual descriptions of the entire image, which helps ground the agent's understanding of the overall environment and the plausibility of certain interactions within it. By dynamically injecting this external knowledge, the framework ensures that predictions are semantically consistent with both the visual evidence and real-world knowledge.

\begin{figure*}[t]
  \begin{center}
    \centerline{\includegraphics[width=\textwidth]{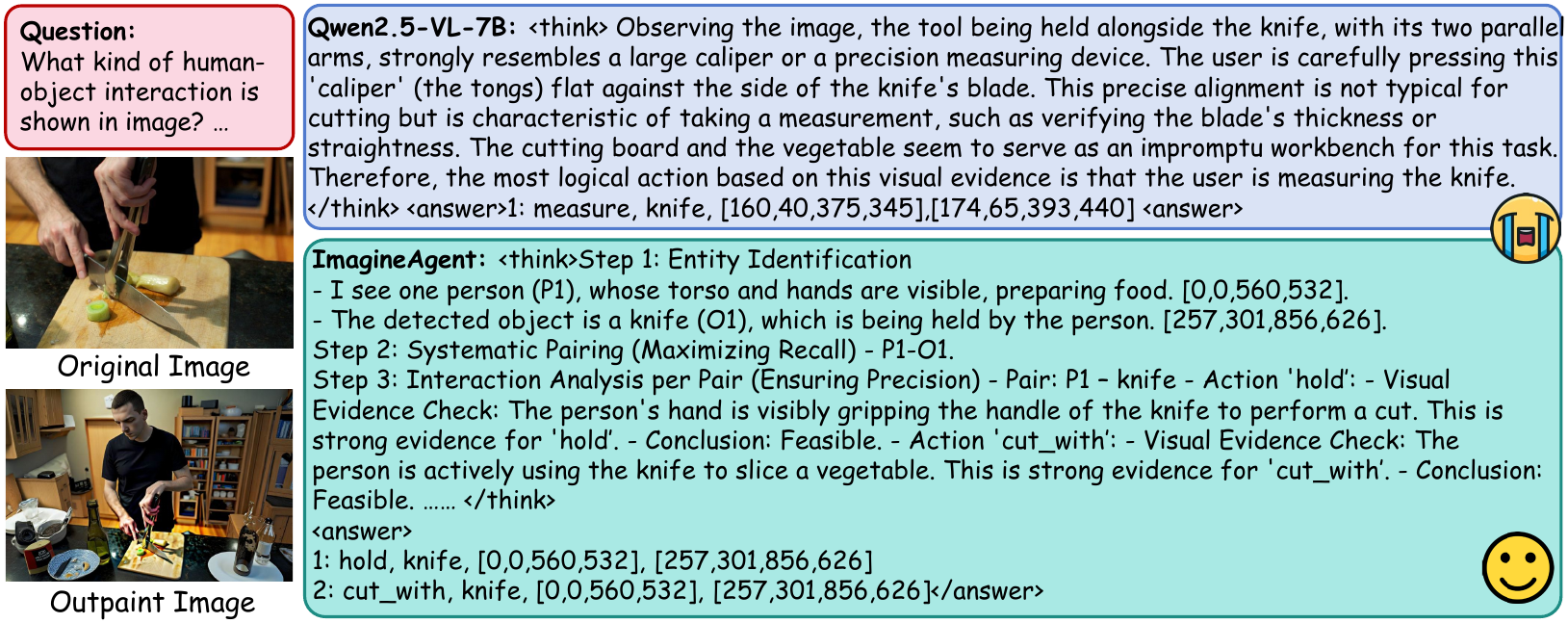}}
    \caption{
    \textbf{Case Study between Qwen2.5-VL-7B and our method.} Our method can accurately identify comprehensive HOI predictions, while Qwen2.5-VL-7B misclassifies the interaction labels.
    }
    \vspace{-0.2in}
    \label{fig:case3}
  \end{center}
\end{figure*}

\begin{figure*}[t]
  \begin{center}
    \centerline{\includegraphics[width=\textwidth]{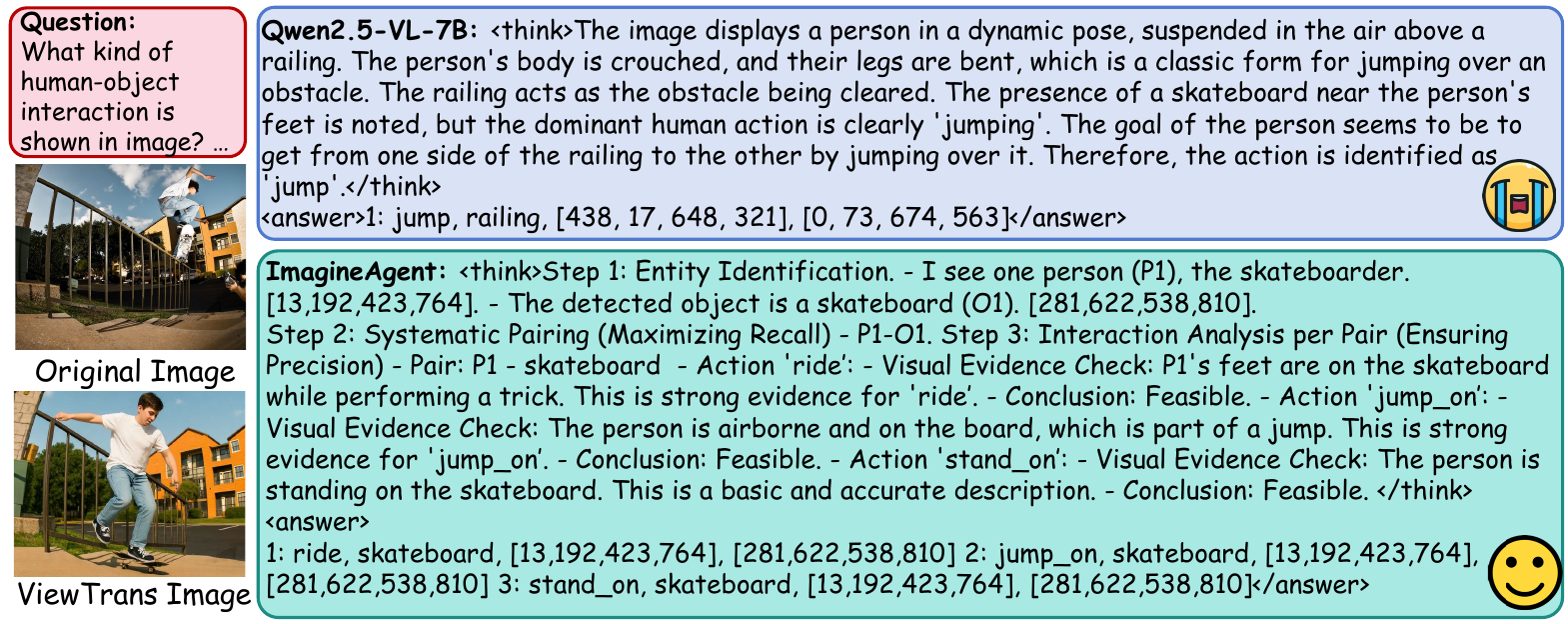}}
    \caption{
    \textbf{Case Study between Qwen2.5-VL-7B and our method.} Our method can accurately identify comprehensive HOI predictions, while Qwen2.5-VL-7B misclassifies the interaction labels.
    }
    \vspace{-0.2in}
    \label{fig:case2}
  \end{center}
\end{figure*}

\subsection{Generative Imagination Tools}
\label{Generation}
\begin{figure}[ht]
  \begin{center}
    \centerline{\includegraphics[width=\columnwidth]{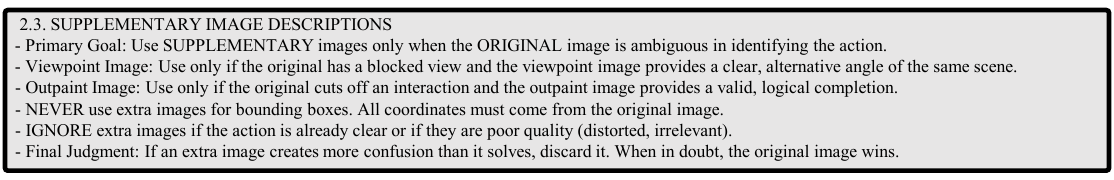}}
    \caption{
      Detailed prompt of supplementary image descriptions.
    }
    \label{prompt0}
    \vspace{-0.2in}
  \end{center}
\end{figure}

To address challenges of occlusion-induced ambiguity, our agent is endowed with generative imagination capabilities through the \textbf{BAGEL} model~\citep{Bagel_2025_arxiv}. This toolset operates in two distinct modes: \textbf{Outpaint} ($T_o$) and \textbf{Viewpoint Transform} ($T_v$). The Outpaint function is employed when key contextual elements are cut off by the image frame, allowing the agent to reconstruct a more complete scene and better infer the overall activity. The Viewpoint Transform function is selected when an object or part of an interaction is occluded from the current perspective. By synthesizing a plausible alternative view, the agent can "see" behind obstructions, effectively filling in missing visual information. This process provides the agent with an imagined reality upon which to base its reasoning, transforming an incomplete observation into a more comprehensive one and enabling robust decision-making under visual uncertainty. Moreover, to control for potential generative artifacts and ensure the reliability of the agent's reasoning, we embed a strict protocol within its prompt, governing the use of supplementary images. The complete instructions are detailed in Fig.~\ref{prompt0}.
 
\section{Discussion on Generative Hallucination}
\label{sec:hallucination}

ImagineAgent leverages generative tools such as Viewpoint Transform and Outpainting to recover occluded or ambiguous interactions. While these tools improve the semantic clarity of interactions, they inherently carry a risk of producing hallucinated visual content. It is therefore important to contextualize this risk within our task objective: OV-HOI detection fundamentally aims to recognize semantic $\langle \text{human, object, action} \rangle$ triplets rather than to reconstruct precise physical grounding or millimeter-level contact points. Consequently, as long as the core interaction semantics are preserved, positional or geometric deviations in generated views do not compromise recognition accuracy.

This rationale is illustrated in Fig.~\ref{fig:case2}. The original image suffers from severe occlusion of the skateboard, which causes the baseline Qwen2.5-VL-7B to hallucinate incorrect actions. By contrast, the ViewTrans output reconstructs a clearer perspective of the person-skateboard interaction. For HOI detection, the decisive evidence lies in identifying the ``skateboarding'' relational pattern rather than verifying exact foot-to-surface coordinates. Additional results are provided in Fig.~\ref{case1}  and Fig.~\ref{fig:case3}.

To quantify the impact of generative hallucinations, we conducted two complementary analyses:

\begin{enumerate}[leftmargin=*]
    \item \textbf{Empirical Evaluation of Hallucination Frequency.} Using 9,546 HICO-DET test images, we define hallucination conservatively: a prediction is considered hallucinated if the model’s output set of actions and objects does not fully cover the ground-truth categories. Under this metric, 38.7\% of images exhibit some form of hallucination, with the majority (29.4\%) arising from incomplete interaction coverage and only 13.9\% due to incorrect predictions.
    \item \textbf{Semantic Fidelity Verification of Generated Views.} We randomly sampled 600 generated images and evaluated them using both human annotators and an automated verification pipeline based on Qwen-VL-Max. Results indicate a human validation accuracy of 86.7\% and a model-based validation accuracy of 88.0\%, confirming that the majority of generated views maintain high semantic fidelity and provide actionable evidence for interaction recognition.
\end{enumerate}

Our pipeline further mitigates hallucination risk through three key design choices:

\begin{itemize}[leftmargin=*]
    \item \textit{Interaction Filtering}: The agent first detects candidate objects and pre-filters only those with clear interaction potential, ensuring distant or non-interacting objects do not enter the reasoning space.
    \item \textit{Targeted Generative Reasoning}: Generative tools are applied exclusively to pre-filtered interaction pairs, serving as auxiliary signals to recover occluded or ambiguous cues without influencing unrelated objects.
    \item \textit{Reward-Guided Tool Usage}: The agent’s composite reward credits tool usage only when it improves HOI predictions. Invalid or unnecessary tool invocations are penalized, preventing over-reliance on potentially hallucinated content.
\end{itemize}

Despite these safeguards, we acknowledge that generative bias remains an intrinsic limitation of diffusion-based image manipulation. Extreme occlusions, crowded scenes, or severe motion blur may still yield unreliable synthesized cues. Nevertheless, empirical evidence indicates that the semantic recovery provided by generative modules consistently outweighs geometric deviations. This is further supported by the ablation study in Tab.~\ref{tab:ablation_tool} , where disabling the Viewpoint Transform module leads to a consistent performance drop, confirming its effectiveness in resolving visual ambiguities.

In conclusion, while hallucination risk cannot be completely eliminated, our pipeline design, empirical validation, and reward-guided constraints collectively ensure that ImagineAgent relies predominantly on semantically meaningful evidence. This establishes both the robustness and reliability of generative-assisted perception for OV-HOI detection, while transparently acknowledging the remaining limitations for future work.

\section{Prompts}
\begin{figure}[t]
  \begin{center}
    \centerline{\includegraphics[width=\columnwidth]{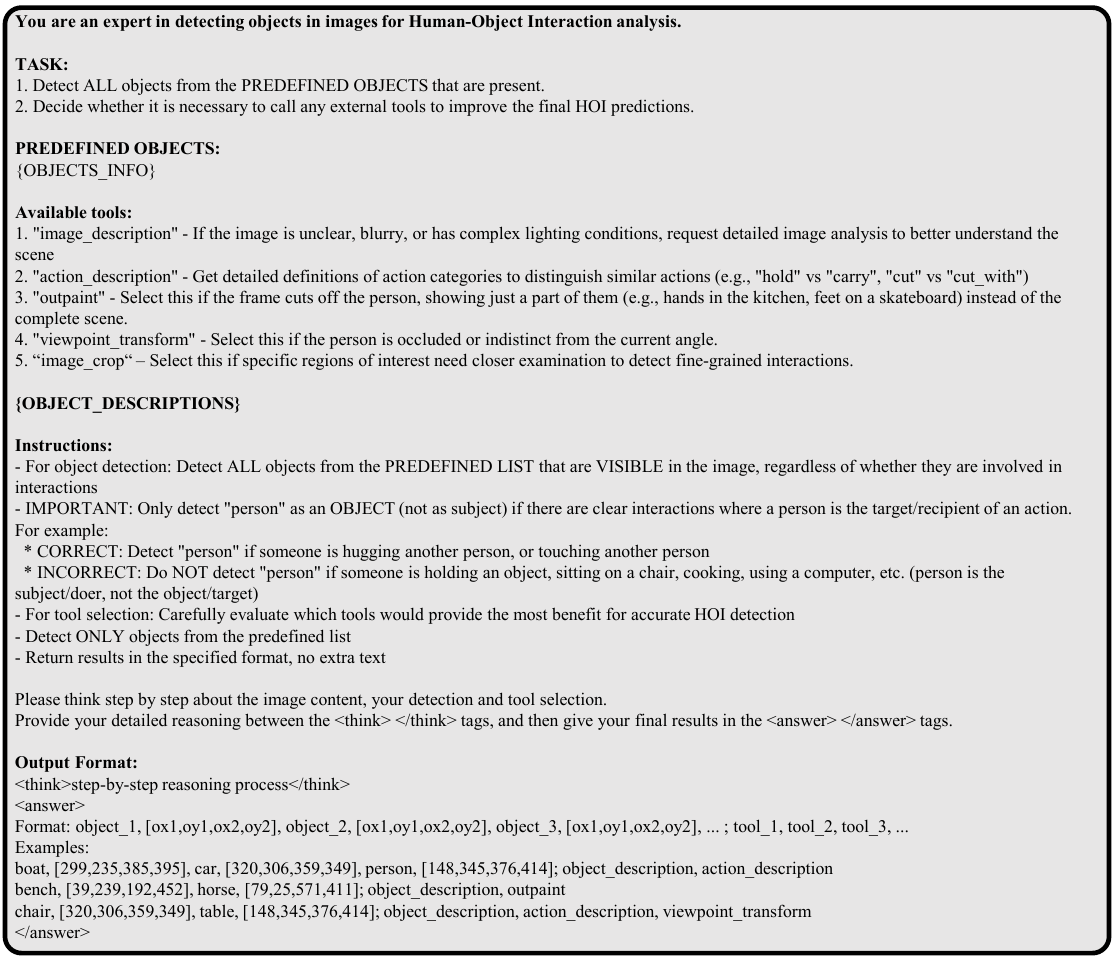}}
    \caption{
      Detailed prompt template in the first round.
    }
    \label{prompt1}
    \vspace{-0.2in}
  \end{center}
\end{figure}

To generate the high-quality, structured reasoning chains essential for our training pipeline, we employ a meticulously designed two-turn prompting strategy. As shown in Fig.~\ref{prompt1}, in the first turn, the model is tasked with initial perception and strategic planning. It is prompted to identify all relevant objects within the image from a predefined vocabulary and to select appropriate tools from a library including image cropping, online retrieval for scene and action explanations, and generative modeling including outpainting and viewpoint transformation. 

\begin{figure}[t]
  \begin{center}
    \centerline{\includegraphics[width=\columnwidth]{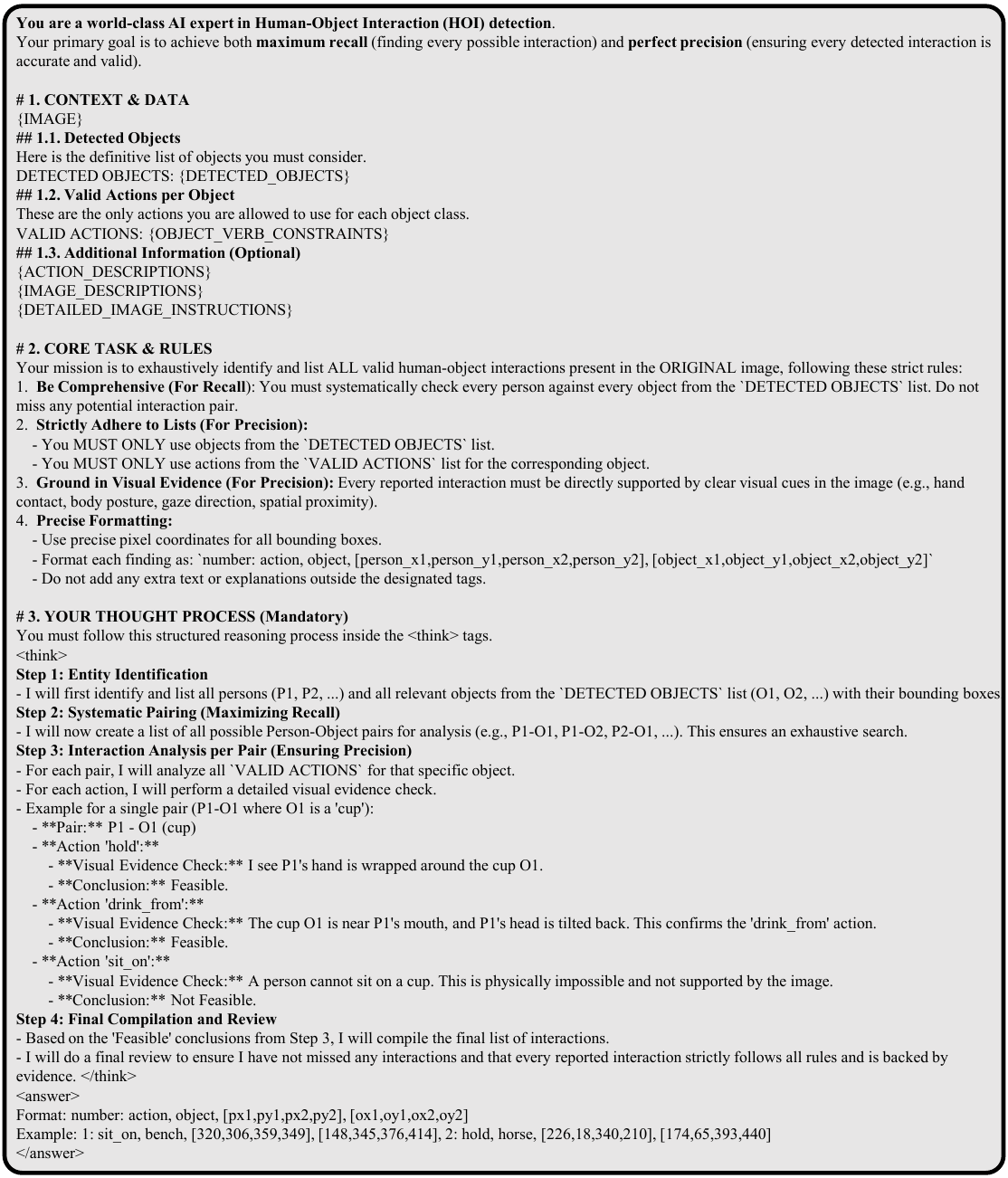}}
    \caption{
      Detailed prompt template in the second round.
    }
    \label{prompt2}
    \vspace{-0.3in}
  \end{center}
\end{figure}

As shown in Fig.~\ref{prompt2}, the second turn focuses on detailed cognitive reasoning and evidence integration. It takes the set of detected objects and any augmented visual or textual data from the deployed tools to generate a constrained, step-by-step analysis. This involves systematically pairing all human and object instances, evaluating a predefined set of valid verbs for each pair, and grounding each potential interaction in direct visual evidence. 

\section{Limitation}
Despite its effectiveness, ImagineAgent still has several limitations. 
First, our framework relies on external tools, including retrieval, image cropping, and generative image editing, which inevitably introduces additional inference cost compared with a single-pass MLLM baseline. 
Although the tool-usage reward encourages effective tool invocation, reducing unnecessary tool calls and further improving inference efficiency remain important directions for future work. 
Second, our current experiments are conducted on HICO-DET and SWIG-HOI, and the proposed training data construction process is tailored to image-based OV-HOI detection. 
Extending the framework to more complex scenarios, such as video-based HOI, embodied interaction understanding, and real-time robotic perception, may require additional temporal reasoning modules and more efficient tool scheduling strategies. 
Finally, our reward design relies on rule-based matching with predefined thresholds for semantic similarity and spatial IoU, more adaptive reward formulations may further improve robustness across datasets with different annotation styles.

\section*{Licenses}
All existing datasets and models used in this paper are credited to their original creators. 
We follow the licenses and terms of use specified in the corresponding original papers, official repositories, or official model cards. 
These assets are used only for research purposes, including OV-HOI evaluation, reasoning-chain construction, semantic matching, and auxiliary tool-augmented inference.

\section*{Impact Statement}

This work aims to improve open-vocabulary human-object interaction understanding, which may benefit assistive perception, robotics, and embodied AI systems. However, HOI recognition also has potential risks in surveillance or activity-monitoring applications, especially when used without consent or human oversight. In addition, the generative tools used in our framework, such as Outpaint and Viewpoint Transform, may have dual-use concerns if misused to synthesize misleading visual evidence. Our method is intended for research use, and deployment in sensitive scenarios should follow appropriate ethical, legal, and safety constraints.


\newpage
\section*{NeurIPS Paper Checklist}

\begin{enumerate}

\item {\bf Claims}
    \item[] Question: Do the main claims made in the abstract and introduction accurately reflect the paper's contributions and scope?
    \item[] Answer: \answerYes{} 
    \item[] Justification: We emphasize the contributions and scope in the Introduction.

\item {\bf Limitations}
    \item[] Question: Does the paper discuss the limitations of the work performed by the authors?
    \item[] Answer: \answerYes{} 
    \item[] Justification: The limitation of the proposed algorithm has been discussed in the supplementary material.

\item {\bf Theory assumptions and proofs}
    \item[] Question: For each theoretical result, does the paper provide the full set of assumptions and a complete (and correct) proof?
    \item[] Answer: \answerNA{} 
    \item[] Justification: There is no theoretical result.

    \item {\bf Experimental result reproducibility}
    \item[] Question: Does the paper fully disclose all the information needed to reproduce the main experimental results of the paper to the extent that it affects the main claims and/or conclusions of the paper (regardless of whether the code and data are provided or not)?
    \item[] Answer: \answerYes{} 
    \item[] Justification: We provide comprehensive implementation details both in main paper and in supplementary material.

\item {\bf Open access to data and code}
    \item[] Question: Does the paper provide open access to the data and code, with sufficient instructions to faithfully reproduce the main experimental results, as described in supplemental material?
    \item[] Answer: \answerNo{} 
    \item[] Justification: As we promised, the data and code will be released upon the publication of our paper.

\item {\bf Experimental setting/details}
    \item[] Question: Does the paper specify all the training and test details (e.g., data splits, hyperparameters, how they were chosen, type of optimizer) necessary to understand the results?
    \item[] Answer: \answerYes{} 
    \item[] Justification:  The experimental setup, including data splits, training and testing detailed, are provided in Method and Experiments sections.

\item {\bf Experiment statistical significance}
    \item[] Question: Does the paper report error bars suitably and correctly defined or other appropriate information about the statistical significance of the experiments?
    \item[] Answer: \answerNo{} 
    \item[] Justification: We follow the default evaluations in the HOI detection field, which doesn’t require error bars.

\item {\bf Experiments compute resources}
    \item[] Question: For each experiment, does the paper provide sufficient information on the computer resources (type of compute workers, memory, time of execution) needed to reproduce the experiments?
    \item[] Answer: \answerYes{} 
    \item[] Justification: We provide them in implementation details of main paper and supplementary material.
    
\item {\bf Code of ethics}
    \item[] Question: Does the research conducted in the paper conform, in every respect, with the NeurIPS Code of Ethics \url{https://neurips.cc/public/EthicsGuidelines}?
    \item[] Answer: \answerYes{} 
    \item[] Justification: This work conforms the NeurIPS Code of Ethics.

\item {\bf Broader impacts}
    \item[] Question: Does the paper discuss both potential positive societal impacts and negative societal impacts of the work performed?
    \item[] Answer: \answerYes{} 
    \item[] Justification: We provide a substantive impact statement in the supplementary material, discussing both positive applications of OV-HOI understanding, such as assistive perception and robotics, and potential risks, including surveillance misuse, erroneous human-object interaction predictions, and dual-use concerns introduced by generative tools such as Outpaint and Viewpoint Transform.
    
\item {\bf Safeguards}
    \item[] Question: Does the paper describe safeguards that have been put in place for responsible release of data or models that have a high risk for misuse (e.g., pre-trained language models, image generators, or scraped datasets)?
    \item[] Answer: \answerNA{} 
    \item[] Justification: The proposed method uses pre-trained models. This proposed methods is safe under the safeguards of adopted pre-trained models.

\item {\bf Licenses for existing assets}
    \item[] Question: Are the creators or original owners of assets (e.g., code, data, models), used in the paper, properly credited and are the license and terms of use explicitly mentioned and properly respected?
    \item[] Answer: \answerYes{}
    \item[] Justification: We credit the original creators of all existing datasets, models, and tools used in this work, including HICO-DET, SWIG-HOI, Qwen-VL/Qwen3-VL, CLIP, and BAGEL. We follow the licenses and terms of use specified in their original papers, official repositories, or official model cards, and use these assets only for research purposes.

\item {\bf New assets}
    \item[] Question: Are new assets introduced in the paper well documented and is the documentation provided alongside the assets?
    \item[] Answer: \answerYes{}{} 
    \item[] Justification: There is no new assets released in this work.
    \item[] Justification: This work introduces \emph{hicodet-6K}, a curated reasoning dataset for OV-HOI supervised fine-tuning. We document its construction pipeline, filtering and validation procedure, and data usage in Appendix. The current submission does not publicly release the dataset for anonymity, and we will clarify the release status upon publication.

\item {\bf Crowdsourcing and research with human subjects}
    \item[] Question: For crowdsourcing experiments and research with human subjects, does the paper include the full text of instructions given to participants and screenshots, if applicable, as well as details about compensation (if any)? 
    \item[] Answer: \answerNA{} 
    \item[] Justification: The paper does not involve crowdsourcing nor research with human subjects.

\item {\bf Institutional review board (IRB) approvals or equivalent for research with human subjects}
    \item[] Question: Does the paper describe potential risks incurred by study participants, whether such risks were disclosed to the subjects, and whether Institutional Review Board (IRB) approvals (or an equivalent approval/review based on the requirements of your country or institution) were obtained?
    \item[] Answer: \answerNA{} 
    \item[] Justification: There is no research with human subjects in this work.

\item {\bf Declaration of LLM usage}
    \item[] Question: Does the paper describe the usage of LLMs if it is an important, original, or non-standard component of the core methods in this research? Note that if the LLM is used only for writing, editing, or formatting purposes and does \emph{not} impact the core methodology, scientific rigor, or originality of the research, declaration is not required.
    \item[] Answer: \answerNA{} 
    \item[] Justification: LLM does not impact the core methodology, scientific rigorousness, or originality of the research.

\end{enumerate}

\end{document}